%% file: main.tex
\RequirePackage[l2tabu, orthodox]{nag}
\documentclass[12pt]{article}

\usepackage[parfill]{parskip}
\usepackage[margin=1in]{geometry}
\usepackage[defaultlines=3,all]{nowidow}

\usepackage[T1]{fontenc}
\usepackage{microtype}
\usepackage[english]{babel}
\usepackage{libertine}             
\usepackage{titlesec}
\usepackage{nicefrac}
\titleformat*{\section}{\Large\bfseries}

\usepackage{amsmath}
\usepackage{amssymb}
\usepackage{amsthm}
\usepackage{mathtools}

\usepackage[dvipsnames]{xcolor}

\usepackage[sort]{natbib}
\usepackage{bibunits}

\usepackage{hyperref}
\hypersetup{colorlinks,linktoc=all}
\usepackage[all]{hypcap}
\hypersetup{citecolor=Violet}
\hypersetup{linkcolor=black}
\hypersetup{urlcolor=MidnightBlue}

\usepackage{algorithm}
\usepackage{algorithmic}
\usepackage{xcolor}

\newcommand{\fgpxl}{FGPX-Shapley-L }
\newcommand{\fgpxg}{FGPX-Shapley-G }

\newtheorem{Theorem}{Theorem}
\newtheorem{Definition}[Theorem]{Definition}

\newtheorem{Lemma}[Theorem]{Lemma}

\newtheorem{Proposition}[Theorem]{Proposition}

\newcommand{\cS}{\mathcal{S}}

\newcommand{\cT}{\mathcal{T}}
\newcommand{\cD}{\mathcal{D}}
\newcommand{\cZ}{\mathcal{Z}}
\newcommand{\GP}{\mathcal{GP}}
\newcommand{\bE}{\mathbb{E}}
\newcommand{\w}[1]{w_{|\mathcal{#1}|}}
\newcommand{\Cov}{\mathrm{Cov}}

\newcommand{\mobius}{M\"{o}bius }
\newcommand{\V}{\mathbb{V}}
\newcommand{\X}{\textbf{X}}
\newcommand{\M}{\textbf{M}}
\newcommand{\tk}{\tilde{k}}
\newcommand{\bL}{\mathbf{L}}
\newcommand{\R}{\mathbb{R}}
\newcommand{\x}{\pmb x}
\newcommand{\y}{\pmb y}
\newcommand{\z}{\pmb z}
\newcommand{\alphab}{\pmb \alpha}
\newcommand{\onevec}{\pmb 1}
\newcommand{\postdata}{\X, \y}

\title{Exact Shapley Attributions in Quadratic-time for FANOVA Gaussian Processes}

\author{
    Majid Mohammadi\textsuperscript{\rm 1,2} \and 
    Krikamol Muandet\textsuperscript{\rm 2} \and 
    Ilaria Tiddi\textsuperscript{\rm 1} \and 
    Annette Ten Teije\textsuperscript{\rm 1} \and  
    Siu Lun Chau\textsuperscript{\rm 3}
}
\date{
\footnotesize{
    \textsuperscript{\rm 1}Department of Computer Science, Vrije Universiteit Amsterdam, The Netherlands\\
    \textsuperscript{\rm 2}Rational Intelligence Lab, CISPA Helmholtz Center for Information Security, Germany\\
    \textsuperscript{\rm 3}College of Computing \& Data Science, Nanyang Technological University, Singapore\\[2ex]}
    \large \today
}

\begin{document}

\maketitle

\begin{abstract}
    Shapley values are widely recognized as a principled method for attributing importance to input features in machine learning. However, the exact computation of Shapley values scales exponentially with the number of features, severely limiting the practical application of this powerful approach. The challenge is further compounded when the predictive model is probabilistic---as in Gaussian processes (GPs)---where the outputs are random variables rather than point estimates, necessitating additional computational effort in modeling higher-order moments. In this work, we demonstrate that for an important class of GPs known as FANOVA GP, which explicitly models all main effects and interactions, exact Shapley attributions for both local and global explanations can be computed in \emph{quadratic} time.
    For \emph{local, instance-wise explanations}, we define a stochastic cooperative game over function components and compute the \emph{exact stochastic Shapley value} in quadratic time only, capturing both the expected contribution and uncertainty. For \emph{global explanations}, we introduce a deterministic, variance-based value function and compute exact Shapley values that quantify each feature’s contribution to the model’s overall sensitivity. Our methods leverage a closed-form (stochastic) \mobius representation of the FANOVA decomposition and introduce recursive algorithms, inspired by Newton's identities, to efficiently compute the mean and variance of Shapley values. Our work enhances the utility of explainable AI, as demonstrated by empirical studies, by providing more scalable, axiomatically sound, and uncertainty-aware explanations for predictions generated by structured probabilistic models. 
\end{abstract}



\input{sections/01_introduction}

\input{sections/02_background}

\input{sections/03_method_local}

\input{sections/04_method_global}
\input{sections/06_experiments}

\input{sections/07_discussion}

\input{sections/10_acknowledgement}

\newpage\clearpage
\bibliographystyle{unsrtnat}
\bibliography{ref}

\newpage\clearpage

\newpage \clearpage
\appendix
\input{sections/08_appendix}

\end{document}

%% file: sections/01_introduction.tex
\section{Introduction}
As machine learning (ML) systems are increasingly deployed in high-stakes applications, the demand for interpretability has grown substantially. Practitioners and regulators alike now seek models whose predictions can be understood and trusted---not only globally, across the entire data distribution, but also locally, for specific predictions. To address this need, the literature offers two distinct approaches: (i) designing inherently interpretable models, such as linear models or generalized additive models (GAMs), or (ii) applying post-hoc interpretation methods like SHAP~\cite{shap} or LIME~\cite{lime} to interpret complex, black-box models. However, interpretability is context-dependent. A model that appears interpretable to ML practitioners—such as GAM—may remain opaque to domain experts or end-users, such as medical professionals. In such cases, even inherently interpretable models may require an additional explanatory layer to translate their insights into signals accessible to non-technical stakeholders.

\emph{Functional ANOVA Gaussian Processes} (FANOVA GPs)~\citep{orthogonal_kernel} are a class of probabilistic models that combine the flexibility of GPs with the interpretability of functional ANOVA decompositions. They represent the prediction function as a sum of functions defined over subsets of input features, where each term models either a main effect or a higher-order interaction. By enforcing functional ANOVA constraints~\citep{FANOVA}, these components become orthogonal (see Section \ref{sec: prelims}), allowing the contribution from each subset to be uniquely and meaningfully identified. Unlike conventional kernels, which typically entangle all features through a single joint interaction term, FANOVA GPs encode a structured hierarchy of interactions---from individual features to the full feature set---making the decomposition both interpretable and expressive. This structure enables precise attribution of predictive behavior to specific feature sets, while preserving GP's nonparametric nature and the ability to quantify uncertainty. As a result, FANOVA GPs offer a compelling foundation for interpretable probabilistic modeling.

Nonetheless, as FANOVA GPs impose an interpretable additive structure by construction, the number of potential interactions grows exponentially with the number of input features. Consequently, identifying the overall contribution of each feature---especially in the presence of higher-order interactions---can become cognitively intractable for human users, much like interpreting a random forest by examining each tree's decision logic. Moreover, existing interpretation methods for FANOVA GP either lack axiomatic support or are computationally inefficient. For example, global sensitivity analysis techniques based on first-order Sobol indices~\citep{sobol, AGP_ortho} ignore interaction effects, fail to satisfy key axioms such as efficiency~\citep{sobol_shapley}, and often underestimate feature importance. On the other hand, although the GPSHAP algorithm of \citet{gp-shap} could, in principle, be applied to FANOVA GPs, it does not exploit their additive structure---a structure that, as we show, enables significant computational savings. Moreover, the standard expectation-based value functions they employ require additional estimation, introducing potential sources of error. In contrast, our approach adopts a more natural value function tailored to FANOVA models~\citep{shapley_coopGames,mohammadi2025computing}, allowing us to compute exact Shapley values without estimation, thereby avoiding approximation errors entirely. Our approach also yields stochastic explanations that account for the predictive uncertainty inherent in GPs.


In short, this paper proposes algorithms that leverage the structure of FANOVA GPs to deliver both \emph{local} and \emph{global} explanations via the Shapley value~\citep{shapley}, computed \emph{exactly} and in \emph{quadratic time}. Our key technical insight is that the orthogonal additive decomposition allows us to derive a closed-form (stochastic) \mobius representation of the prediction function, enabling efficient computation of Shapley values and their associated uncertainty propagated from the GPs. We develop recursive algorithms, inspired by Newton's identities, that avoid exponential enumeration over feature subsets and compute the Shapley values for local and global explanation efficiently. The contributions of this paper can be summarized as follows:
\begin{enumerate}
    \item \textbf{Explaining the uncertain.} We propose the \textit{FGPX-Shapley-L} (\textbf{F}ANOVA \textbf{GP}-based e\textbf{X}act \textbf{Shapley} value for \textbf{L}ocal explanation) algorithm to provide local explanations in the form of stochastic Shapley values~\citep{ma2008shapley} for FANOVA GPs that can be computed in \emph{quadratic} time through a recursive algorithm using Newton's identities.
    \item \textbf{Explaining the uncertainty.} We propose the FGPX-Shapley-G (G for global) algorithm to attribute the variance of the prediction as global feature attribution, following the spirit in classical global sensitive analysis literature. A recursive, quadratic-time algorithm is also devised to ensure computational efficiency.
\end{enumerate}

Taken together, our results show that interpretability, axiomatic attribution, and computational efficiency need not be at odds---provided that the model admits the right structure. FANOVA GPs, and the methods we develop to explain them, offer a new blueprint for scalable and rigorous explainable machine learning.

%% file: sections/02_background.tex
\section{Preliminaries}
\label{sec: prelims}
\paragraph{Notations.} We denote the set of $d$ features by $\cD$, and its power set $2^{\cD}$. The training set $\{\x_i, y_i \}_{i=1}^n$ consists of $\x_i \in \R^d$ and $y\in \R$ (regression) or $y\in \{1,\dots, \ell\}$ ($\ell$-class classification). Let $\X\in \R^{n\times d}$ denote the full input dataset, and $\X_{\cS}$ its restriction to feature subset $\cS \in 2^{\cD}$; the corresponding sample space is $X_\cS$. The probability density of feature $i$ is denoted $p(X_i)$.
We use calligraphic letters for sets, capital letters for random variables, and bold-faced lower- and upper-cased letters for vectors and matrices, respectively. Element-wise product is denoted by $\odot$, expectation over data and model $f$'s predictive distribution is denoted by $\bE_{X}$ and $\bE_f$, and variances by $\V_X$ and $\V_f$. 


\subsection{FANOVA Gaussian process}
We focus our exposition on regression tasks for clarity. Nonetheless, the proposed Shapley algorithms remain applicable to classification problems, provided that the posterior distribution of the FANOVA GP is available.

We assume observations $y$ arise from a latent function $f(\x)$ corrupted by Gaussian noise, and we place a Gaussian process prior on $f$. Following \citet{AGP}, we enforce an additive structure on $f$:
\begin{equation}
f(\x)\;=\;\sum_{\cS\subseteq\cD}f_{\cS}(\x_{\cS}),
\label{eq:func-decompos}
\end{equation}
where each $f_{\cS}(\x_{\cS})$ depends only on the feature subset $\cS$ of input $\x$.  This additive formulation, referred to as the \emph{functional decomposition} of $f$, enables the model to capture main effects and interactions of arbitrary order in a structured way. This structure is induced via an additive kernel:
\begin{align}\label{eq:additive kernel}
k^{\mathrm{add}_q}(\x,\x') &= \sigma_q^2\sum_{1\leq i_1 \leq \cdots \leq i_d \leq d} \left[\prod_{l=1}^q k_{i_l} (x_{i_l},x'_{i_l}) \right],\, 
\end{align}
which assigns scale parameter $\sigma_q^2$ to all $q$-way interactions. The actual GP is then built with the full additive kernel by summing over interaction orders
$
k^{\mathrm{add}}(\x,\x')
=\sum_{q=0}^d k^{\mathrm{add}_q}(\x,\x'),
$
with $k^{\mathrm{add_0}}=\sigma_0^2$. A GP with such a kernel is referred to as \emph{additive Gaussian process (AGP)}. Although evaluating $k^{\mathrm{add}}$ involves $\binom{d}{q}$ terms per order $q$, \citet{AGP} showed that a recursion based on Newton's identities yields a polynomial‐time algorithm, making the approach practical. 

In short, by placing a GP prior $\mathcal{GP}(0, k^{\mathrm{add}})$ over the latent function $f$, with observation model $y=f + \epsilon$ where independent noise \( \epsilon \sim \mathcal{N}(0, \sigma_n^2) \), the predictive posterior over \( y_{\x} \) at a new input \( \x \) is again a GP, i.e. $y_{\x}\mid \X,\y \sim
\GP \bigl(\xi,\kappa\bigr),$
with the mean and covariance functions being expressed as:
\begin{align}\label{eq:gp-var}
&\xi(\x)=k^{\mathrm{add}}(\x,\X)^{\top}
\alphab, \,\, \alphab = \Sigma^{-1}\y, \,\, \Sigma = k^{\mathrm{add}}(\X,\X)+\sigma_n^2 I, \nonumber \\
&\kappa(\x,\x')
=k^{\mathrm{add}}(\x,\x')-
k^{\mathrm{add}}(\x,\X)^{\!\!\top}
\Sigma^{-1}
k^{\mathrm{add}}(\x',\X).
\end{align}


One challenge with AGPs is identifiability, as many basis functions can sum to $ f(\x) $. \citet{orthogonal_kernel} addressed this using the functional ANOVA decomposition \citep{FANOVA} that imposes two key constraints: (i) each component satisfies the zero mean condition, $ \mathbb{E}_{X_{\cS}}[f_{\cS}(\x_{\cS})] = 0 $ for every non-empty subset $ \cS $; and (ii) components are ``mutually orthogonal'' in the sense that $ \mathbb{E}_X[f_{\cS}(\x_{\cS}) f_{\cS'}(\x_{\cS'})] = 0 $ for any $ \cS \neq \cS' $. \citet{orthogonal_kernel} put forward a class of kernels that satisfy the above conditions. In particular, for any base kernel $k_i$, the \textit{constrained kernel} \( \tilde{k}_i \) is defined as:
\begin{equation}\label{eq:orthogonal kernel}
    \tilde{k}_i(x_i, x'_i) = k_i(x_i, x'_i) - \frac{\int k_i(x_i, s) p(s) ds \int k_i(x_i', s) p(s) ds}{\int \int k_i(s, t) p(s)p(t)dsdt},
\end{equation}
with $p$ some density defined over the data space.
The higher-order kernels $\tilde{k}^{add_q}$ are then constructed by multiplying corresponding $\tilde{k}_i$ as in equation \eqref{eq:additive kernel}, and the additive constrained kernel is defined as $$\tilde{k}^{add}(\x,\x') = \sum_{q=0}^d \tilde{k}^{add_q}(\x,\x').$$ \citet{orthogonal_kernel} showed that functions drawn from a GP with this kernel satisfy the ANOVA decomposition conditions. As a follow-up,
\citet{AGP_ortho} demonstrated for several popular kernels, such as the squared exponential and categorical kernels, with a Gaussian density over features, $\tilde{k}_i$ admits an analytical expression; whereas for other densities or kernels, the integration in equation \eqref{eq:orthogonal kernel} could be estimated by the empirical probability measure based on the training samples. For the rest of the paper, we refer to the additive GP satisfying the FANOVA conditions as \textit{FANOVA GP}.

\subsection{(Stochastic) Shapley values}
The Shapley value~(SV)~\citep{shapley} is a solution concept from cooperative game theory that provides an axiomatic framework for fairly distributing the total value generated by a group back to its individual members. Its appeal lies in satisfying a unique set of desirable properties: efficiency, symmetry, dummy, and linearity. Formally, given a tuple \((\cD, v )\), where \(\cD\) is the set of $d$ players and \(v: 2^\cD \to \R \) is a real-valued set function, the SV for player \(i \in \cD\) is defined as a specific weighted average of their marginal contributions across all possible coalitions. The function $v(\cS)$ can be interpreted as the “worth” or value generated by the subset $\cS$.

However, in many settings—such as probabilistic modeling or decision-making under uncertainty—the value function may not be deterministically specified, but rather known only up to a distribution. In such cases, it is natural to ask whether an analogue of the Shapley value exists. Indeed, it does. \citet{ma2008shapley} and \citet{gp-shap} formalized the notion of stochastic value functions \(\nu: 2^\cD \to \mathcal{L}(\mathbb{R})\), which assign to each coalition $\cS$ a real-valued probability distribution, making $\nu(\cS)$ a real-valued random variable, thereby capturing the uncertainty in the value or importance attributed to that subset. This extension gives rise to the stochastic Shapley value~(SSV), which generalizes the classical formulation to settings where importance must be assessed in a probabilistic rather than deterministic manner. Formally, given player set $\cD$ and stochastic value function $\nu$, the SSV of player $i$ is given by
\begin{align}\label{eq:ssv}
\phi_i(\nu) = \sum_{\substack{\cS \subseteq \cD \setminus \{i\}}} c_{|\cS|} \Big( \nu(\cS \cup \{i\}) - \nu(\cS) \Big),
\end{align}
where $c_{|\cS|} = \frac{|\cS|!(d - |\cS| - 1)!}{d!}$. This formula looks analogous to the classical Shapley value, which is not surprising as the classical value function is a special case of the stochastic counterpart (e.g. use a Dirac function), but note that $\phi_i(\nu)$ is now a random variable. A key advantage of the stochastic formulation is that it naturally allows one to compute higher-order uncertainty statistics—most notably, the variance $\mathbb{V}(\phi_i(\nu))$, which quantifies uncertainty in the resulting attributions themselves.

\paragraph{A novel stochastic \mobius representation.} We extend the theory of stochastic cooperative games by introducing the stochastic \mobius representation, a concept that has not yet been explored in the existing literature. We first state our results formally:

\begin{Definition}[Stochastic \mobius representation]
    Given a stochastic value function $\nu$, the stochastic \mobius representation $\mu:2^\cD\to\mathcal{L}(\mathbb{R})$ is defined as:
    \begin{align*}
        \mu(\cT) = \sum_{\cS \subseteq \cT} (-1)^{|\cT|-|\cS|} \nu(\cS).
    \end{align*}
\end{Definition}
The quantity \(\mu(\cT)\) can be interpreted as the stochastic \mobius coefficient of coalition \(\cT\), representing its unique contribution to the stochastic value function. This representation yields the following result:
\begin{Proposition}\label{prop:var phi}
    Given stochastic cooperative game $\nu$ and its stochastic \mobius representation $\mu$, we have
    \begin{itemize}
        \item The stochastic Shapley value for player $i\in\cD$ is $$\phi_i(\nu) = \sum_{\cT \subseteq \cD, \cT \ni i} |\cT|^{-1} \mu(\cT)$$, and
        \item the variance of $\phi_i(\nu)$ naturally follows as $$\V\big(\phi_i(\nu)\big) = \sum_{\substack{\cT \subseteq \cD \\ \cT \ni i}} \sum_{\substack{\cT' \subseteq \cD \\ \cT' \ni i}} \tfrac{1}{|\cT| |\cT'|} \Cov\big(\mu(\cT), \mu(\cT')\big).$$
    \end{itemize}
\end{Proposition}
While the utility of this representation may not be immediately evident, we demonstrate in the next section that adopting the stochastic \mobius perspective leads to significant computational advantages. In particular, it enables efficient evaluation of both the mean and variance of the SSV.

%% file: sections/03_method_local.tex
\section{Explaining the locally uncertain with stochastic Shapley values}

\paragraph{Shapley value for local explanation.}  Owing to its desirable axiomatic properties, the Shapley value has garnered significant attention as a principled method for feature attribution in machine learning~\citep{shap, gemfix, svsvl}. In the context of local explanations, a common approach treats input features as players in a cooperative game and defines a value function tailored to a given predictive model $g$ and input \( \x \). A widely used formulation is
$$v(\cS) = \bE[g(X) \mid X_{\cS} = \x_{\cS}] - \bE[g(X)],$$
which quantifies the importance of a feature subset \( \cS \subseteq \cD \) by measuring the change in the model’s expected output when features in \( \cD \setminus \cS \) are marginalized out. Intuitively, this reflects the added predictive value of observing \( X_{\cS} = \x_{\cS} \) compared to having no feature information. 

\paragraph{The GP-SHAP algorithm.} To extend this to GPs, where predictions are probabilistic, \citet{gp-shap} generalized the value function to a stochastic setting and proposed GP-SHAP. They showed that the conditional expectation of a GP remains stochastic and used the theory of conditional mean processes~\citep{chau_bayesimp_2021,chau_deconditional_2021,chau2022rkhs} to characterize the resulting stochastic value function analytically. While their approach applies to FANOVA GPs, it does not exploit the model’s additive structure and relies on standard approximation techniques such as those used in Kernel SHAP~\citep{shap}. Moreover, the conditional expectation-based value function incurs estimation error. In contrast, our approach advocates a different value function tailored to models with functional decomposition. This structure enables exact, estimation-free computation of both the value function and the stochastic Shapley values. Crucially, it also reduces the computational complexity of computing exact SSVs from exponential to quadratic time.

\paragraph{Stochastic Shapley values for FANOVA GP.} When a predictive model $f$ admits a functional decomposition, \citet{shapley_coopGames} and \citet{mohammadi2025computing} proposed an alternative value function for measuring subset contributions that aligns more closely with the sensitivity analysis literature. Specifically, for a decomposable $f$, an input $\x$, the (stochastic) value function $\nu$ is defined as:
$$\nu_{\x}(\cS) = \sum_{\cT \subseteq \cS} f_\cT(\x_{\cT}).$$
That is, the value of a subset \( \cS \subseteq \cD \) is computed by summing all component functions whose indices are contained within \( \cS \), thereby capturing the total contribution of features in \( \cS \) to the overall prediction. This natural choice of value function not only preserves the tractability of GP models but also enables the development of a quadratic-time algorithm for computing Shapley values.


\begin{Proposition}\label{prop:gp value func}
Given a posterior FANOVA GP $p(f\mid \postdata)$ defined in equation~\eqref{eq:gp-var}, an input $\x$, the value function $\nu_{\x}$ defined above, we have that:
\begin{itemize}
    \item $\nu_{\x}$ is a GP over $2^{\cD}$ with mean and covariance functions:
    \begin{align*}
    &\xi_{\nu_{\x}}(\cS | \X, \y) = \sum_{\cT \subseteq \cS} \sigma_{|\cT|}^2  \tk_{\cT}(\x_{\cT},\X_{\cT})^\top \alphab, \cr
    &\kappa_{\nu_{\x}}(\cS, \cS' | \X, \y) = \sum_{\cT \subseteq \cS \cap \cS'} \sigma_{|\cT|}^2  \tk_{\cT}(\x_{\cT}, \x_{\cT}) \cr 
    &\,- \sum_{\cT \subseteq \cS} \sum_{\cT' \subseteq \cS'} \sigma^2_{|\cT|}\sigma^2_{|\cT'|}\tk_{\cT}(\x_{\cT},\X_{\cT})^\top \Sigma^{-1} \tk_{\cT'}(\x_{\cT'},\X_{\cT'}).
    \end{align*}
    \item the \mobius representation $\mu_{\x}$ is also a GP over $2^{\cD}$ with mean and covariance functions:
    \begin{align*}
    &\xi_{\mu_{\x}}(\cS | \postdata) = \sigma_{|\cS|}^2  \tk_{\cS}(\x_{\cS},\X_{\cS})^\top \alphab, \cr
    &\kappa_{\mu_{\x}}(\cS, \cS' | \postdata) = \delta_{\cS\cS'} \sigma^2_{|\cS|}\tk_{\cS}(\x_{\cS}, \x_{\cS}) \cr 
    & \qquad - \sigma^2_{|\cS|}\sigma^2_{|\cS'|}\tk_{\cS}(\x_{\cS},\X_{\cS})^\top \Sigma^{-1} \tk_{\cS'}(\x_{\cS'},\X_{\cS'}),
    \end{align*}
    where $\delta_{\cS\cS'} = 1$ when $\cS=\cS'$, and $0$ otherwise. 
\end{itemize}
\end{Proposition}
As Proposition~\ref{prop:gp value func} demonstrates, the \mobius representation $\mu_{\x}$ is neater to work with compared to $\nu_{\x}$. In fact, the analytical expressions and recursive algorithms for computing the mean and variance of the SSV for FANOVA GP are also more straightforward to derive from $\mu_{\x}$ than $\nu_{\x}$. We present the analytical expression of the resulting SSV under $\nu_{\x}$ in the following proposition.

\begin{Proposition}\label{prop:gp sv}
    Let $w_s = \nicefrac{\sigma_s^2}{s}$ for some scaler $s$. The stochastic Shapley values $\pmb \phi(\nu_{\x})$ follows a multivariate Gaussian distribution with mean $\xi_{\pmb\phi}\in \R^d$ and covariance matrix $\mathbf{K}_{\pmb \phi} \in \R^{d\times d}$, where
    \begin{align*}
        \xi_{\pmb\phi}= \sum_{\cS\subseteq \cD, \cS \ni i}w_{|\cS|} \tk_{\cS}(\x_{\cS},\X_{\cS})^\top \alphab,
    \end{align*}
    and $[\mathbf{K}_{\pmb\phi}]_{i,j} = \operatorname{Cov}(\phi_i(\nu_{\x}), \phi_j(\nu_{\x}))$ can be expressed as 
    \begin{align*}
        &\sum_{\cS\subseteq\cD, \cS \ni i}\sum_{\cT\subseteq\cD, \cT \ni j}
   \delta_{\cS\cT}\tfrac{\sigma^2_{|\cS|}}{|\cS|^2} \tk_{\cS}(\x_{\cS},\x_{\cS})  - \Big(\sum_{\substack{\cS\subseteq\cD \\ \cS \ni i}} \w{S}\tk_{\cS}(\x_{\cS},\X_{\cS})\,\Big)\Sigma^{-1}\, \Big(\sum_{\substack{\cT\subseteq\cD\\ \cT \ni j}} \w{T} \tk_{\cT}(\X_{\cT},\x_{\cT})\Big).
    \end{align*}
\end{Proposition}
 


The following lemma shows how the efficiency axiom of stochastic SV relates to the posterior mean function. 
\begin{Lemma}\label{lem:local additivity}
For a fixed input $\x$, let $\{\phi_j(\nu_{\x})\}_{j=1}^d$ be the stochastic Shapley values of the
FANOVA\,GP and denote their posterior means by
$\xi_{\phi_j}(\x)=\bE[\phi_j(\x)\mid\postdata]$.
Let $\xi(\x)=\bE[f(\x)\mid\postdata]$ be the GP posterior mean and
$\xi_\varnothing=\bE[f_{\varnothing}\mid\postdata] = \sigma_0\alphab^\top\onevec$ the posterior mean of the constant
component.
Then,
$
\sum_{j=1}^{d}\xi_{\phi_j}(\x) = \xi(\x)- \xi_\varnothing.
$
\end{Lemma}

\subsection{The FGPX-Shapley-L algorithm}

In the following,  we introduce one of our main contributions: the FGPX-Shapley-L (\textbf{F}ANOVA \textbf{GP}-based e\textbf{X}act \textbf{Shapley} value for \textbf{L}ocal explanation) algorithm, which relies on the following recursive formulation.

\paragraph{Quadratic-time recursive computation.}
A key challenge in computing the SSV is its exponential complexity: both the mean and variance involve summations over exponentially many subsets of features, making exact computation intractable for large \(d\). The following theorem introduces a recursive formulation that enables computing the \textit{exact} Shapley value in quadratic time, significantly reducing the computational burden. The recursion is based on Newton’s identities—originally used to efficiently construct additive kernels—and here adapted for computing SSVs. Since the technical details of these identities (involving elementary symmetric polynomials and power sums) can be intricate, we defer their full definition and derivation to Appendix~\ref{apx:newton identities}. For the main text, we encourage readers to treat these constructions abstractly: they provide a principled way to summarize interaction terms compactly and recursively, allowing us to scale up exact computations without explicitly enumerating all feature subsets. We begin by presenting the recursion for computing the mean of the Shapley value.

\begin{Theorem}
\label{th:mean-var recursion}
Let $\z_j := \tilde{k}_j(x_j, \X_j)$ and define the set $\cZ_{-i} = \{\,\z_j : j \in \cD \setminus \{i\} \}$.
Let the \emph{elementary symmetric polynomials} (ESPs) over $\cZ_{-i}$ be defined recursively as:
$$  
e_r(\cZ_{-i}) = \frac{1}{r} \sum_{s=1}^{r} (-1)^{s-1} e_{r-s}(\cZ_{-i}) \odot p_s(\cZ_{-i}),
$$ 
with $e_0(\cZ_{-i}) = \onevec,$
where \( p_s(\cZ_{-i}) = \sum_{\z \in \cZ_{-i}} \z^s \) is the element-wise power sum.  
Define the intermediate vector:
\begin{align}\label{eq:l recursion}
\pmb \ell_i := \z_i \odot \sum_{q=0}^{d-1} w_{q+1} e_q(\cZ_{-i}).
\end{align}
\begin{enumerate}
    \item  
    The mean of the SSV for feature $i$ is $\xi_{\phi_i} = \pmb\ell_i^\top \alphab.$
    
    \item The variance of the SSV for feature $i$ is given by:
    \begin{align}\label{eq:var recursive}
    \V_f(\phi_i \mid \postdata) =
    \bigg(\bar{z}_i \sum_{q=0}^{d-1} \tfrac{\sigma^2_{q+1}}{(q+1)^2} \, e_{q}(\bar{\cZ}_{-i}) \bigg)
    - \pmb\ell_i^\top \Sigma^{-1} \pmb\ell_i,
    \end{align}
    where $\bar{z}_j := \tilde{k}_j(x_j, x_j)$, $\bar{\cZ}_{-i} = \{\,\bar{z}_j : j \in \cD \setminus \{i\} \}$, and $e_q(\bar{\cZ}_{-i})$ is the ESPs of order $q$ for set \( \bar{\cZ}_{-i} \).
\end{enumerate}
\end{Theorem}

The intuition to arrive at the recursion in equation \eqref{eq:l recursion} is to use the additive structure of the kernel function in FANOVA GP to factorize $\tk_{\cS}(\x_{\cS}, \X_{\cS})$, bring $\tk_j(x_j,\X_j)$ out of the summation, and write the remaining part of the summation as the weighted ESPs. This recursion is used to simplify the exact computation of the mean and variance of SSVs, and reduces the computational complexity from exponential to quadratic in the number of features. This represents a substantial improvement in scalability. Using the recursion in equation~\eqref{eq:l recursion}, the mean of SSVs can be computed directly. The variance requires an additional recursion, which we derive in equation~\eqref{eq:var recursive}. The time complexity of computing SSVs for a test instance using our method is $O(nd^2)$. Specifically, for each feature $i$, we remove its corresponding kernel vector to compute ESPs over the remaining $d{-}1$ features, which incurs a cost of $O(nd^2)$. For the variances of SSVs, the total cost remains $O(nd^2)$, since the recursion in the first term is of $O(d^2)$ complexity, and the second term is computed from $\pmb \ell_i$'s with $O(nd^2)$ complexity, leading to an overall $O(nd^2)$ complexity.

To compute the full covariance structure of SSVs, a closely related recursive formulation with a similar complexity can be employed. The complete algorithm, along with a numerically stable implementation of the ESPs, is provided in Appendix~\ref{apx:ssv-efficient}. In the next section, we introduce a recursive algorithm for computing variance-based Shapley values, aimed at providing global explanations.

%% file: sections/04_method_global.tex
\section{Explaining the global uncertainty with Shapley values}

Besides local explanation, it is also important to quantify global feature importance to provide a holistic view of the overall predictive performance.
A common approach to provide this is through global sensitivity analysis, where we compute the ratio of the variance corresponding to a feature set and the total variance. \citet{orthogonal_kernel,AGP_ortho} introduced the first-order Sobol index for FANOVA GP and provided an analytical solution under some conditions. However, as \citep{sobol_shapley} already argued, the first-order Sobol index only captures individual contributions and neglects interactions between features, limiting its ability to fully represent the importance of features in complex models. 
Ironically, in our case, since FANOVA GPs explicitly model feature interactions, relying on first-order Sobol indices that ignore these interactions is inappropriate.
A possibility is to compute higher-order Sobol index for each feature set, but that becomes immediately impractical due to the exponential number of components. 

Realizing these shortcomings, \citet{sobol_shapley} advocated the use of Shapley values instead. Before we introduce the corresponding value function, we recall that FANOVA GP enjoys the following variance decomposition:
\begin{align}\label{eq:variance decomposition}
    \V_X(f(\x)) = \sum_{\cS \subseteq \cD} \V_X(f_{\cS}(\x_{\cS})).
\end{align}
Analogous to the local explanation setting, we define a variance-based value function \(v : 2^{\cD} \to \R \) over feature subsets as:
$$
v_G(\cS) := \sum_{\cT \subseteq \cS} \V_X(f_{\cT}(\x_{\cT})),
$$
which quantifies the total contribution of the feature subset \(\cS\) to the output variance. This value function admits a \mobius representation \(m_G : 2^{\cD} \to \R \), given by
\(
m_G(\cS) = \V_X\big(f_{\cS}(\x_{\cS})\big),
\)
where each \mobius coefficient represents the variance attributable to interaction subset \(\cS\). Given this setup, we now present a closed-form expression for the global Shapley value \(\phi_i(v_G)\) under the posterior of the FANOVA GP model, which allows for efficient, recursive computation.

\begin{Theorem}\label{th:sv global}
Let $\alphab$ be the posterior mean as in Equation~\ref{eq:gp-var}. Then, the global Shapley value of feature $i$ based on the variance-based sensitivity analysis, shown by $\phi_i(v_G)$, is
\begin{align}\label{eq:global sv}
    \phi_i(v_G) := \sum_{\substack{\cS \subseteq \cD \\ \cS \ni i}} \tfrac{m_G(\cS)}{|\cS|}= \pmb\alpha^\top \bigg( \sum_{\substack{\cS \subseteq \cD \\ \cS \ni i}} \tfrac{\sigma_{|\cS|}^4}{|\cS|} \bL_{\cS} \bigg) \pmb\alpha,
\end{align}
where $\bL_{\cS} = \odot_{i\in {\cS}} \int_{\mathcal{X}_i} \tk_i(x_i, \X_i) \tk_i(x_i, \X_i)^\top d p(x_i)$.
\end{Theorem}

For all $i\in\cD$, $\bL_i$ can also be estimated with an empirical measure and proven to have an analytical solution with a Gaussian density of features \citep{AGP_ortho}.  Note that the Shapley value in equation \eqref{eq:global sv} is still exponentially expensive to compute. We now present an efficient quadratic-time computation based on this equation.

\begin{Theorem}\label{th:global recursion}
Let $\cZ_{-i} = \{\bL_j : j \in \cD \setminus \{i\} \}$, $p_s(\cZ_{-i}) = \sum_{\bL \in \cZ_{-i}} \bL^{s}$ denote the element-wise power-sum matrices, and define the ESPs recursively via Newton’s identities:
$$
e_0(\cZ_{-i}) = \onevec,
e_r(\cZ_{-i}) = \frac{1}{r} \sum_{s=1}^{r} (-1)^{s-1} e_{r-s}(\cZ_{-i}) \odot p_s(\cZ_{-i}).
$$
Then, the matrix
$$
\M_i = \sum_{\cS \ni i} \frac{\sigma_{|\cS|}^4}{|\cS|} \,\bL_{\cS}
$$
admits the closed-form recursion
$$
\M_i = \bL_i \odot \sum_{r=0}^{d-1} \frac{\sigma_{r+1}^4}{r+1} \, e_r(\cZ_{-i}).
$$
Consequently, the global Shapley value for feature $i$ can be computed as
$
\phi_i(v_G) = \alphab^\top \M_i \alphab.
$
\end{Theorem}

Similar to Lemma~\ref{lem:local additivity}, we show how the efficiency axiom of SV relates to the variance of the FANOVA GP.

\begin{Lemma}\label{lem:global additivity}
Let $f(\x)$ be a function drawn from the FANOVA GP with observational noise $y = f(\x) + \epsilon$, where $\epsilon \sim \mathcal{N}(0, \sigma_n^2)$. Then the total output variance decomposes as:
$
\V_X(y) - \sigma_n^2 = \sum_{i=1}^{d} \phi_i(v_G).
$
\end{Lemma}

%% file: sections/06_experiments.tex
\section{Illustrations and Experiments}

We empirically demonstrate our methods through (1) an illustration of stochastic explanations, (2) run-time comparisons, and (3) feature selection problems. We address the \(O(n^3)\) complexity of training GPs through inducing point formulation. We employ the standard squared exponential kernel as our base kernel. Further details on experimental setups, data generation, model tuning, explainers, and feature selectors, along with additional experiments, are provided in Appendix~\ref{apx:experiments}. The experiments were executed on a 24-core machine with 16GB of RAM and an RTX4000 GPU.


\begin{figure}[t]
    \centering
    \includegraphics[width=0.8\linewidth]{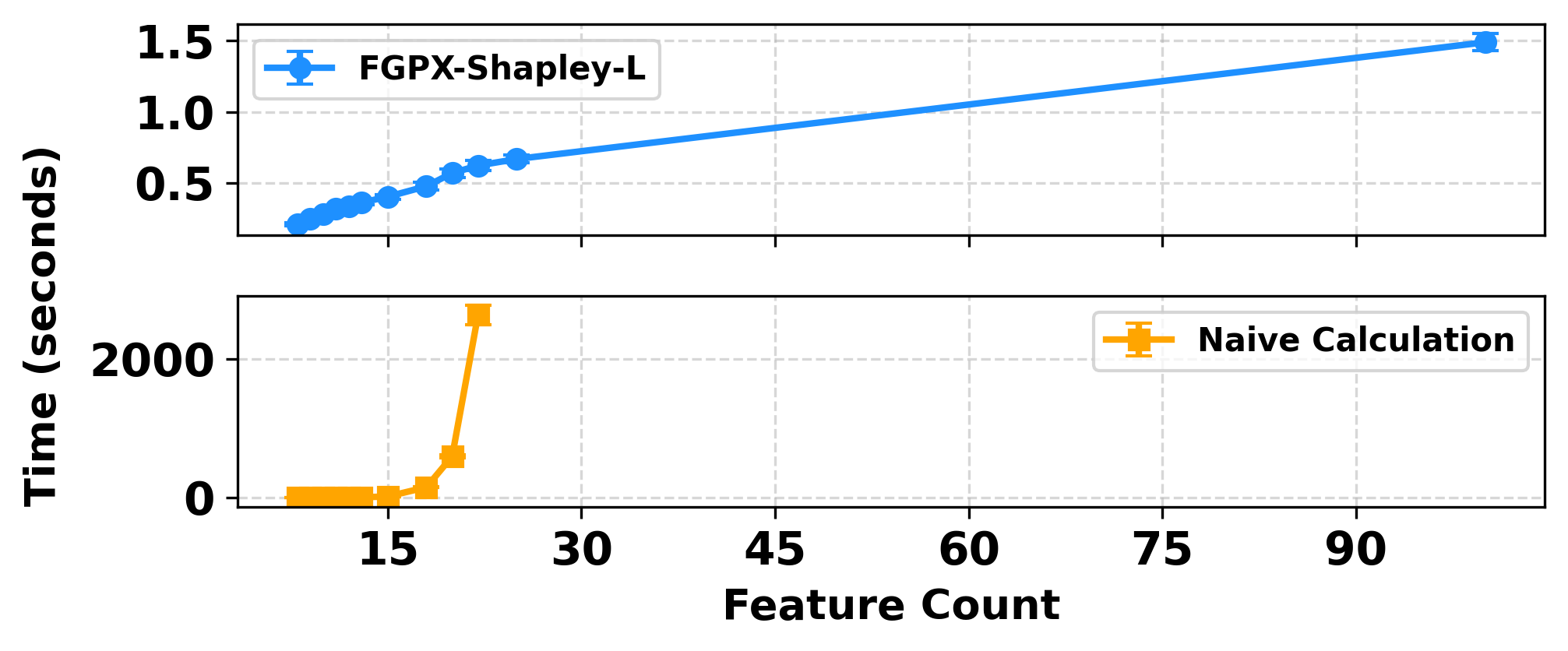}
    \caption{Comparison of execution times between \fgpxl and the n\"{a}ive Shapley computation for different numbers of features, with a maximum time limit of five hours imposed for generating the explanations.}
    \label{fig:time naive}
\end{figure}

\subsection{Time Comparison}
We generated synthetic datasets with 8 to 100 features, each with 500 samples. For each dataset, we trained a FANOVA GP and randomly selected 30 test instances. We then computed SSVs using our method, FGPX-Shapley-L, and compared its average execution time to a naive implementation that computes only the mean of the Shapley values using the same Möbius-based value function. Figure~\ref{fig:time naive} shows execution times across feature dimensions. Up to 12 features, both methods perform similarly (ours: 0.33s vs. naive: 1.35s). Beyond that, the naive method becomes impractical—taking over 2600 seconds at 22 features—while ours stays efficient (0.62s). At 25 features, the naive method failed to complete within 5 hours, but our method scaled to 100 features in just a few seconds. These results clearly demonstrate the scalability and computational advantage of FGPX-Shapley-L for efficient SSV computation.

\subsection{Illustration on how stochastic SV can provide richer explanations}

\begin{figure}[t]
    \centering
\includegraphics[width=0.7\linewidth]{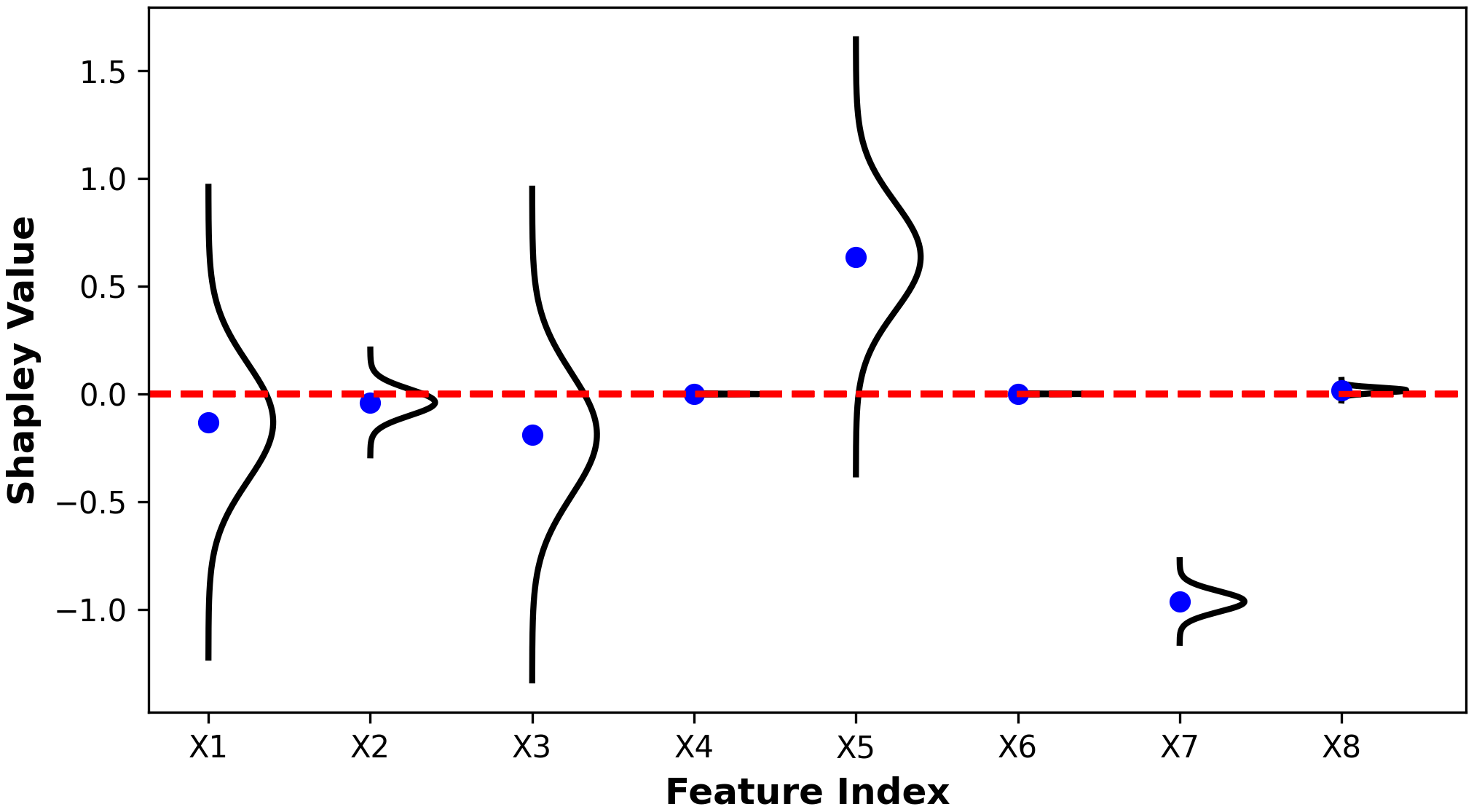}
    \caption{Local SSVs from the energy efficiency dataset. While mean importance is similar for $X1,X2$, the large explanation variance in $X1$ suggests we should calibrate our trust in the model's explanation.}
    \label{fig:illustration}
\end{figure}

At this point, a sensible reader could ask: ``What does uncertainty in explanation bring to the table?''. We address this by providing a demo-use case of our method through the \texttt{energy efficiency} dataset from the UCI repository. The dataset consists of  768 samples and 8 input features describing the architectural and environmental characteristics of buildings. The goal is to predict either the heating or cooling load of a building. We fit a FANOVA GP to the dataset and compute the SSVs for a single, randomly selected test point. Figure~\ref{fig:illustration} shows the resulting SSVs, capturing both the estimated importance and the uncertainty of each feature.

\begin{figure}[t]
    \centering
\includegraphics[width=.65\linewidth]{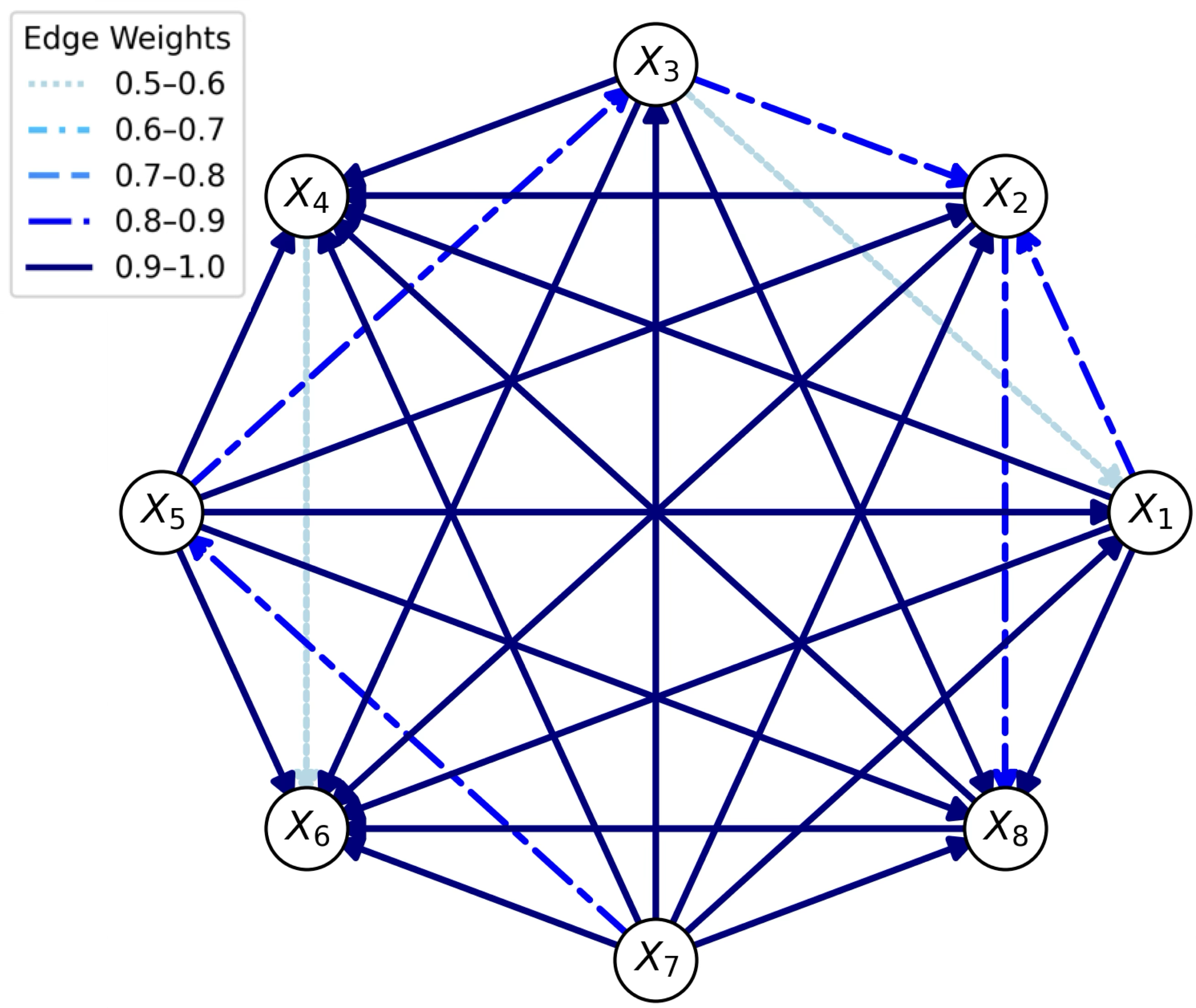}
\caption{Directed graph of pairwise importance comparisons. An edge from feature \(i\) to \(j\) indicates \(\mathbb{P}(|\phi_i| \geq |\phi_j|)\), with edge color and style encoding its strength.}
    \label{fig:illustration_graph}
\end{figure}

SSVs offer a probabilistic view of feature importance, enabling richer analyses than deterministic approaches. Prior work has used their joint Gaussian structure to uncover dependencies between features or interpret acquisition functions in Bayesian optimization \citep{gp-shap,adachi2024looping}. Building on this, we propose a new application of SSVs that allows us to quantify uncertainty when comparing feature importance. The idea is straightforward: given two SSVs, $\phi_i$ and $\phi_j$, a practitioner may wish to assess how likely it is that feature $i$ is more important than feature $j$. This corresponds to estimating the probability $\mathbb{P}(|\phi_j| \geq |\phi_i|)$. Although this quantity is analytically intractable, the joint distribution of $[\phi_i, \phi_j]$ is Gaussian, making it straightforward to estimate via sampling. Figure~\ref{fig:illustration_graph} visualizes the uncertainty-aware comparison of feature importance as a directed graph for an arbitrary instace. We estimate the pairwise probabilities \(\mathbb{P}(|\phi_i| \geq |\phi_j|)\) for all feature pairs using 1,000 samples from the joint distribution of SSVs. Each node represents a feature, and each directed edge reflects the likelihood that one feature is more important than another. The edge color and line style encode the strength of these probabilities, offering a probabilistic view of relative feature importance under uncertainty.

\subsection{Comparing local explanation methods through influential feature recovery}

While feature attribution is inherently an unsupervised learning problem with no nontrivial objective ground truth, it is common to assess its quality via an influential feature recovery experiment. The following presents the evaluation of \fgpxl for local explanations by comparing it in recovering the influential features with several state-of-the-art attribution methods, including Sampling SHAP (S-SHAP) \citep{sampling_shap}, Unbiased SHAP (U-SHAP) \citep{unbiased_shap}, Bivariate SHAP (Bi-SHAP) \citep{bivariateSHAP}, LIME \citep{lime}, and MAPLE \citep{maple}. We also implement a version of Kernel SHAP and GP-SHAP, where the value function in Proposition~\ref{prop:gp value func} is used.


\noindent \paragraph{Synthesized Experiments.} We assess the performance of \fgpxl for feature selection using three synthesized datasets, each containing 20 features and 1,000 samples. These datasets are specifically designed with predefined influential features and varying levels of interaction complexity. Details about the dataset generation process are provided in the appendix.

\begin{figure}[t!]
    \centering
    \includegraphics[width=\linewidth]{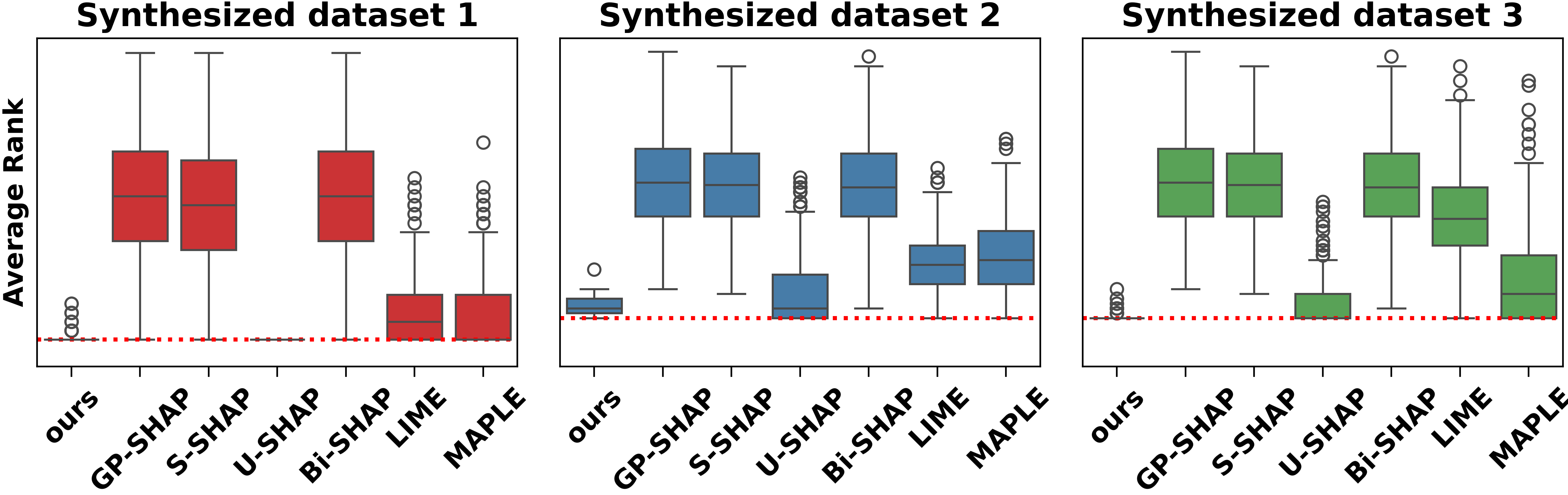}
    \caption{The comparison of explainable methods on four synthesized data sets. }
    \label{fig:exp synthesized ds}
\end{figure}

For each dataset, we randomly select 500 instances and generate explanations using \fgpxl and baseline methods. The evaluation measures how accurately each method ranks the most influential features for individual instances. For each explanation, we rank the features by their assigned importance and compute the average rank of the ground-truth influential features across all instances. Figure~\ref{fig:exp synthesized ds} presents the average rank of the most influential features across 500 selected instances for three synthesized datasets. A lower average rank indicates better identification of influential features. The methodology for computing the average rank is explained in the appendix. The red dotted horizontal line in the figure represents the ideal average rank, corresponding to perfect identification of influential features. The figure demonstrates that \fgpxl consistently delivers robust and accurate local explanations across all datasets. It outperforms other methods by reliably identifying the most influential features averaged over instances, highlighting its effectiveness in providing reliable local explanations.

\begin{figure}[t!]
    \centering
    \includegraphics[width=\linewidth]{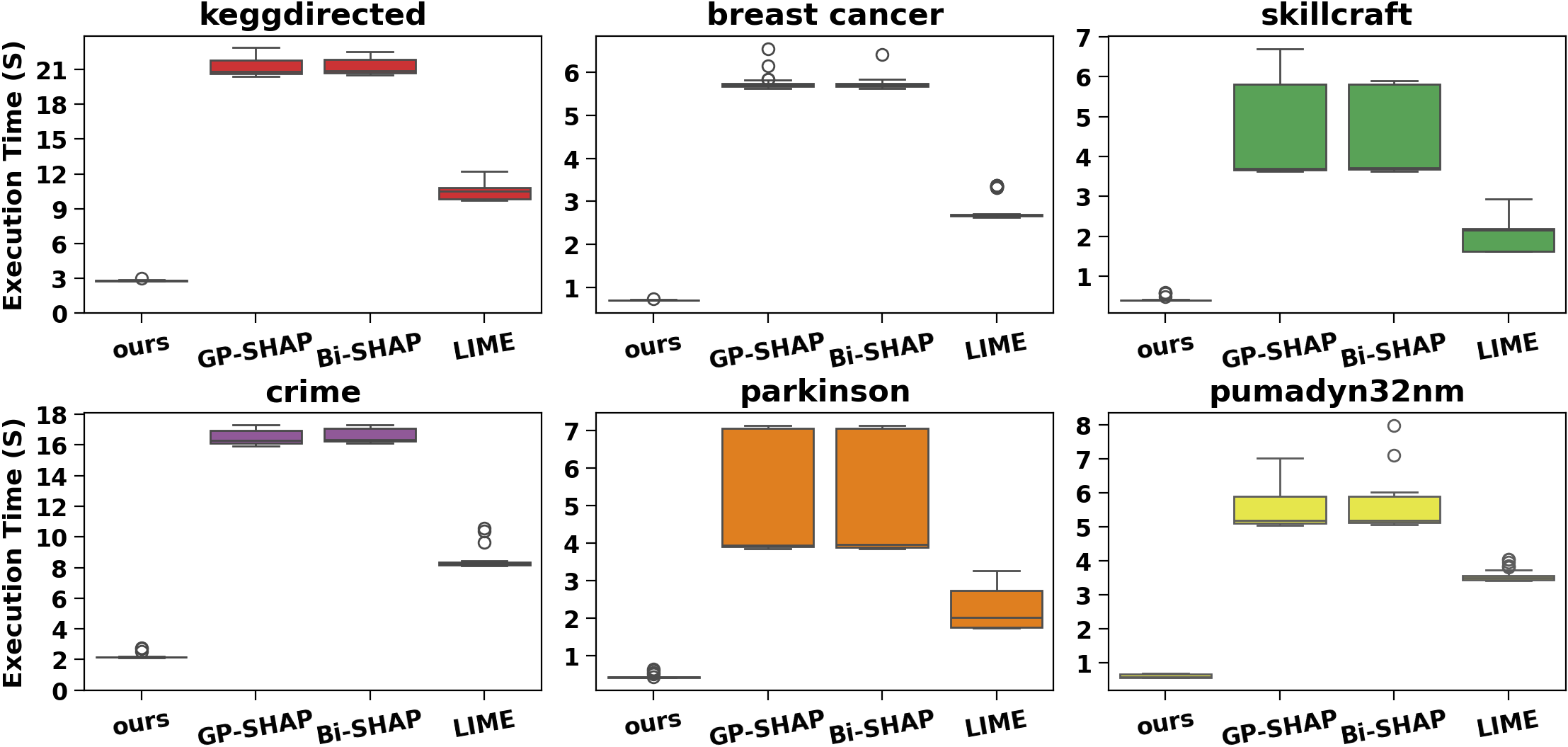}
    \caption{The comparison of different explanations in terms of execution time.}
    \label{fig:time comparison}
\end{figure}

\begin{figure}[t!]
    \centering
    \includegraphics[width=\linewidth]{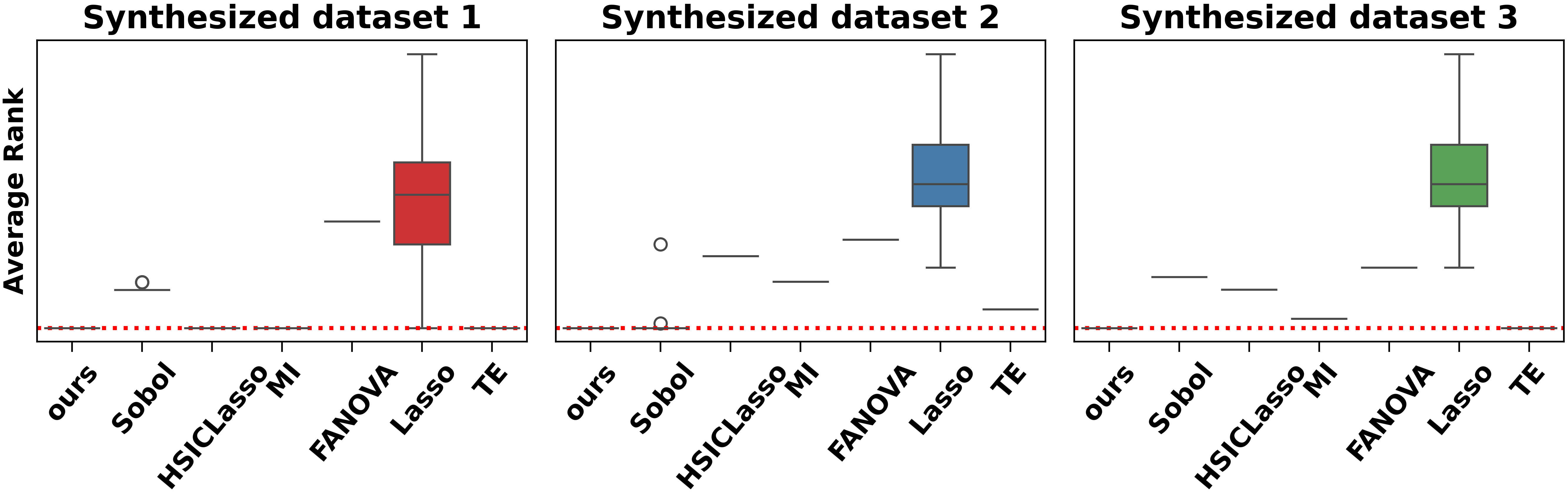}
    \caption{The comparison of feature selectors on four synthesized data sets. The ideal average rank (the lower, the better) is shown by red dotted lines.}
    \label{fig:fs synthesized}
\end{figure}

\noindent \paragraph{Time Comparison.} We assess the execution time of various explainable methods on real datasets. For this evaluation, 50 samples are randomly selected from six real datasets, and different methods are applied to generate local explanations from a trained FANOVA GP. Figure~\ref{fig:time comparison} presents a boxplot showing the average time (in seconds) required to explain an instance from these datasets. Due to the significantly longer execution times of MAPLE and S-SHAP, only the most competitive algorithms are included in the plot for a clearer comparison. A complete plot, including all methods, is provided in the appendix. \fgpxl demonstrates outstanding performance in terms of execution time, significantly outperforming other methods, thanks to the proposed recursive algorithm. More experiments on real datasets are provided in the appendix.

\subsection{Feature Selection with Global Shapley Value}
Similarly, the performance of \fgpxg is compared against several baseline feature selection methods on synthesized and real data sets, including the first-order Sobol index for AGPs \cite{AGP_ortho}, HSICLasso \citep{hsic_lasso}, mutual information (MI) \citep{mi_fs}, Lasso \citep{lasso}, tree ensemble (TE) \cite{tree_ensemble}, and ANOVA F-statistics. 

\noindent \paragraph{Synthesized datasets.}
We first evaluate the performance of \fgpxg on the same synthetic datasets used in the explanation experiments. For each dataset, we compute the average rank of the ground-truth influential features based on the importance scores assigned by each method to the whole dataset. This process is repeated 100 times to ensure statistical reliability. Figure~\ref{fig:fs synthesized} illustrates the average rank of influential features for different methods across three datasets. The results show that \fgpxg consistently outperforms baseline methods, particularly on datasets with complex higher-order interactions and nonlinearity (e.g., dataset 3).

Notably, \fgpxg outperforms the first-order Sobol index, which struggles due to its inability to capture feature interactions. Among the other methods, MI and HSICLasso demonstrate competitive performance in some cases, benefiting from their incorporation of feature interactions via mutual information and kernel functions, respectively. However, \fgpxg exhibits strong and consistent performance across all datasets, making it a reliable choice for feature selection. More experiments on real datasets are provided in the appendix.



%% file: sections/07_discussion.tex
\section{Discussion}
We presented exact algorithms for computing stochastic Shapley values in FANOVA Gaussian processes, enabling both local and global explanations that are axiomatic, uncertainty-aware, and computationally efficient. Our results demonstrate that when the model structure is appropriately designed—such as the orthogonal additive structure of FANOVA GPs—interpretability does not come at the cost of predictive performance. This work highlights that scalable, transparent, and probabilistically grounded explanations are achievable within expressive probabilistic models in quadratic time only.

%% file: sections/10_acknowledgement.tex
\subsection*{Acknowledgement}
This project is funded by the Federal Ministry of Education and Research (BMBF). The DAAD Postdoc-NeT-AI short-term research visit scholarship partially supported the first author. 

%% file: sections/08_appendix.tex
\section{Proofs}

\subsection{Proof of Proposition \ref{prop:var phi}}
The first part of the proposition is obtained by replacing the \mobius representation in the Shapley value formulation. 

For the second part, use the linearity of $\phi_i(\nu)$ in $\mu_\cT(\nu)$ and the definition of covariance:
\[
\V\big(\sum_{\substack{\cT \subseteq \cD \\ \cT \ni i}} \frac{\mu(\cT)}{|\cT|}\big) = \sum_{\substack{\cT \subseteq \cD \\ \cT \ni i}} \sum_{\substack{\cT' \subseteq \cD \\ \cT' \ni i}} \frac{1}{|\cT| |\cT'|} \Cov\big(\mu(\cT), \mu(\cT')\big).
\]

\subsection{Proof of Proposition \ref{prop:gp value func}}
\textbf{Posterior for $\nu_{\x}$:} \\
Since $\nu_{\x}(\cS) = \sum_{\cT \subseteq \cS} f_{\cT}(\x_{\cT})$, the linearity of expectation gives:
\begin{align*}
\xi_{\nu_{\x}}(\cS | \postdata) = \sum_{\cT \subseteq \cS} \bE\bigg(f_{\cT}(\x_{\cT}) | \postdata\bigg) \cr
= \sum_{\cT \subseteq \cS} \sigma_{|\cT|}^2  \tk_{\cT}(\x_{\cT},\X_{\cT})^\top \Sigma^{-1} \y.
\end{align*}
The posterior covariance function, $\forall \cS,\cS' \subseteq \cD$, is:
\begin{align*}
\kappa_{\nu_{\x}}(\cS, \cS' | \postdata) &= \Cov\bigg(\nu_{\x}(\cS), \nu_{\x}(\cS') | \postdata \bigg) \cr 
&= \sum_{\substack{\cT \subseteq \cS \\ \cT' \subseteq \cS'}} \Cov\bigg(f_{\cT}(\x_{\cT}), f_{\cT'}(\x_{\cT'}) | \postdata\bigg).
\end{align*}
To simplify this, we use the prior and posterior covariance. We know that
$
Cov\bigg(f_{\cT}(\x_{\cT}),\,f_{\cT}(\x_{\cT'}) \bigg)
=\delta_{\cT\cT'} \tk_{\cT}(\x_{\cT},\x_{\cT})
$
under the prior. Let $\cT', \cT \subseteq \cD, \cT \neq \cT'$. Under the prior, one can write: 

{\small
\begin{align*}
&\begin{pmatrix}
f_{\cT}(\x_{\cT}) \\
f_{\cT'}(\x_{\cT'}) \\
\y
\end{pmatrix}
\sim \mathcal{N} \bigg(
\begin{pmatrix}
0 \\
0 \\
\mathbf{0}
\end{pmatrix},
\\
&\quad\begin{pmatrix}
\sigma^2_{|\cT|}\tk_{\cT}(\x_{\cT}, \x_{\cT}) & 0 & \sigma^2_{|\cT|}\tk_{\cT}(\x_{\cT}, \X_{\cT}) \\
0 & \sigma^2_{|\cT'|}\tk_{\cT'}(\x_{\cT'}, \x_{\cT'}) & \sigma^2_{|\cT'|}\tk_{\cT'}(\x_{\cT'}, \X_{\cT'}) \\
\sigma^2_{|\cT|}\tk_{\cT}(\X_{\cT}, \x_{\cT}) & \sigma^2_{|\cT'|}\tk_{\cT'}(\X_{\cT'}, \x_{\cT'}) & \Sigma
\end{pmatrix}
\bigg)
\end{align*}
}

We therefore obtain the following under the posterior:
\begin{align}\label{eq:post cov f T S}
&Cov\big(f_{\cT}(\x_{\cT}),f_{\cT'}(\x_{\cT'}) \mid \X, \y \big)
=\delta_{\cT\cT'}\sigma^2_{|\cT|}\tk_{\cT}(\x_{\cT},\x_{\cT}) \cr 
&\, - \sigma^2_{|\cT|}\sigma^2_{|\cT'|}\tk_{\cT}(\x_{\cT}, \X_{\cT})^{\!\top}\Sigma^{-1}\tk_{\cT'}(\x_{\cT'}, \X_{\cT'}).
\end{align}

We can now decompose the covariance over the value functions:
\begin{align*}
&\Cov\bigg(\nu_{\x}(\cS), \nu_{\x}(\cS') | \postdata \bigg) = \underbrace{\sum_{\cT \subseteq \cS \cap \cS'} \sigma^2_{|\cT|}\tk_{\cT}(\x_{\cT}, \x_{\cT})}_{\text{Prior covariance}} \cr 
&- \sum_{\cT \subseteq \cS} \sum_{\cT' \subseteq \cS'} \sigma^2_{|\cT|}\sigma^2_{|\cT'|}\tk_{\cT}(\x_{\cT},\X_{\cT})^\top \Sigma^{-1} \tk_{\cT'}(\x_{\cT'},\X_{\cT'}).
\end{align*}
Finite-dimensional marginals are Gaussian, confirming $\nu_{\x}$ is a GP over $2^{\cD}$.

\textbf{Posterior for $\mu_{\x}$:} \\
Give $\nu_{\x}(\cS) = \sum_{\cT \subseteq \cS} f_{\cT}(\x_{\cT})$, one can readily find that $\mu_{\x}(\cS) = f_{\cS}(\x_{\cS})$. Then, the standard GP regression gives:
\[
\xi_{\mu_{\x}}(\cS | \postdata) = \bE\big( f_{\cS}(\x_\cS) \big) =  \sigma_{|\cS|}^2  \tk_{\cS}(\x_{\cS},\X_{\cS})^\top \Sigma^{-1} \y.
\]
The posterior covariance is obtained from equation~\eqref{eq:post cov f T S} as:
\begin{align*}
&\Cov(\mu_{\x}(\cS), \mu_{\x}(\cS') | \postdata) = \Cov(f_{\cS}(\x_{\cS}), f_{\cT}(\x_{\cT}) | \postdata ) \cr
&\;= \delta_{\cS\cS'} \sigma^2_{|\cS|}\tk_{\cS}(\x_{\cS}, \x_{\cS}) \cr 
& \qquad - \sigma^2_{|\cS|}\sigma^2_{|\cS'|}\tk_{\cS}(\x_{\cS},\X_{\cS})^\top \Sigma^{-1} \tk_{\cS'}(\x_{\cS'},\X_{\cS'}),
\end{align*}
where the cross-term $\sigma^2_{|\cS|}\sigma^2_{|\cS'|}\tk_{\cS}(\x_{\cS},\X_{\cS})^\top \Sigma^{-1} \tk_{\cS'}(\x_{\cS'},\X_{\cS'})$ captures data-induced dependence between $\cS$ and $\cS'$. Finite-dimensional marginals are Gaussian, confirming $\mu_{\x}$ is a GP over $2^{\cD}$.

\subsection{Proof of Proposition \ref{prop:gp sv}}
Since the posterior \mobius components 
$\{\mu_{\x}(\cS)\}_{\cS\subseteq \cD}$ 
are jointly Gaussian, any linear combination of them is again Gaussian.  
In particular, for each feature $i$ and fixed input $\x$,
$
\phi_i(\x)=\sum_{\substack{\cS \subseteq \cD \\ \cS \ni i}}\frac{1}{|\cS|}\,\mu_{\x}(\cS)
$
is Gaussian, where the mean is calculated as:
\begin{align*}
\xi_{\phi_i}(\x)
&=\bE\bigl[\phi_i(\x)\bigr]
=\sum_{\substack{\cS \subseteq \cD \\ \cS \ni i}}\frac{1}{|\cS|}\bE\bigl[\mu_{\x}(\cS)\bigr]\cr 
&=\sum_{\substack{\cS \subseteq \cD \\ \cS \ni i}}\frac{1}{|\cS|}\xi_{\mu}(\cS)
=\sum_{\substack{\cS \subseteq \cD \\ \cS \ni i}}\w{S}
\tk_{\cS}(\x_{\cS},\X_{\cS})^\top\alphab,
\end{align*}

and covariance is computed as
\begin{align*}
&\Cov\bigg(\phi_i(\nu_{\x}),\phi_j(\nu_{\x}) \Bigr| \postdata \bigg) \cr 
&\quad =\Cov\Bigl(\sum_{\substack{\cS \subseteq \cD \\ \cS \ni i}}\frac{1}{|\cS|}\mu_{\x}(\cS),  \sum_{\substack{\cT \subseteq \cD \\ \cT \ni j}}\frac{1}{|\cT|}\mu_{\x}(\cT)\Bigr | \postdata \bigg)\\
&\quad =\sum_{\substack{\cS \subseteq \cD \\ \cS \ni i}}\sum_{\substack{\cT \subseteq \cD \\ \cT \ni j}}
   \frac{1}{|\cS||\cT|}
   \Cov\bigg(\mu_{\x}(\cS),\mu_{\x}(\cT) \Bigr| \postdata \bigg)\\
&\quad =\sum_{\substack{\cS \subseteq \cD \\ \cS \ni i}}\sum_{\substack{\cT \subseteq \cD \\ \cT \ni j}}
   \Bigl[\delta_{\cS\cT}\tfrac{\sigma^2_{|\cS|}}{|\cS|^2} \tk_{\cS}(\x_{\cS},\x_{\cS}) 
     - \w{S}\w{T}\tk_{\cS}(\x_{\cS},\X_{\cS})\,\Sigma^{-1}\, \tk_{\cT}(\X_{\cT},\x_{\cT})
   \Bigr],
\end{align*}
where the last equality follows from equation~\eqref{eq:post cov f T S}. This completes the proof.

Observe that the double sum in the second term factorizes:
\begin{align*}
&\sum_{\substack{\cS, \cT \subseteq \cD \\ \cS \ni i, \cT \ni i}}\w{S}\w{T}
\tk_{\cS}(\x_{\cS},\X_{\cS})\Sigma^{-1} \tk_{\cT}(\X_{\cT},\x_{\cT}) \cr
& =\Bigl(\sum_{\substack{\cS \subseteq \cD \\ \cS \ni i}}\w{S} \tk_{\cS}(\x_{\cS},\X_{\cS})\Bigr)
\Sigma^{-1}
\Bigl(\sum_{\substack{\cT \subseteq \cD \\ \cT \ni j}}\w{T} \tk_{\cT}(\X_{\cT},x_{\cT})\Bigr).
\end{align*}
Combining these two contributions yields the stated result.


\subsection{Proof of Lemma \ref{lem:local additivity}}
Let $\xi_{\mu}(\cS)=\bE\big(f_{\cS}(\x_{\cS})\mid\postdata\big)$ denote the posterior mean of each
additive component.
By construction of the FANOVA\,GP,
\[
\xi(\x)=\xi_{\varnothing}\;+\!\!\sum_{\varnothing\neq\cS\subseteq\cD}\!\!\xi_{\mu}(\cS),
\tag{1}
\]
where $\xi_{\varnothing}$ is the constant ($0$-way) term $\sigma_0 \alphab^\top \onevec$.

For every feature $j$ the posterior mean of its stochastic Shapley value is
\[
\xi_{\phi_j}(\x)
=\sum_{\substack{\cS\subseteq\cD\\ \cS \ni j}} \xi_{\mu_{\x}}(\cS) / |\cS|,
\]
which results in the summation over all Shapley values $j=1,\dots,d$:
\begin{align*}
\sum_{j=1}^{d}\xi_{\phi_j}(\x)
&=\sum_{\varnothing\neq\cS\subseteq\cD} 
\underbrace{\bigl|\cS\bigr|}_{\text{number of $j$ s.t.\ }j\in\cS}\!
\xi_{\mu_{\x}}(\cS) / |\cS| \\
&=\sum_{\varnothing\neq\cS\subseteq\cD}\!\xi_{\mu_{\x}}(\cS).
\end{align*}

It yields:
\[
\sum_{j=1}^{d}\xi_{\phi_j}(\x)
=\xi(\x)-\xi_\varnothing.
\]


\subsection{Proof of Theorem~\ref{th:mean-var recursion}}

We start from the expression for the stochastic Shapley value of feature \(i\), defined via the \mobius representation under the FANOVA GP posterior. As established in Proposition~\ref{prop:gp sv}, the mean and variance of \(\phi_i\) are given by:
\begin{align*}
&\bE_f(\phi_i \mid \postdata) = \sum_{\substack{\cS \subseteq \cD \\ \cS \ni i}} \w{\cS} \cdot \tk_{\cS}(\x_{\cS}, \X_{\cS})^\top \alphab, \\
&\V_f(\phi_i \mid \postdata) = \sum_{\substack{\cS \subseteq \cD \\ \cS \ni i}} \tfrac{\sigma_{|\cS|}^2}{|\cS|^2} \cdot \tk_{\cS}(\x_{\cS}, \x_{\cS})
- \left( \sum_{\substack{\cS \subseteq \cD \\ \cS \ni i}} \w{S} \cdot \tk_{\cS}(\x_{\cS}, \X_{\cS}) \right)^{\!\!\top}
\Sigma^{-1}
\left( \sum_{\substack{\cT \subseteq \cD \\ \cT \ni i}} \w{T} \cdot \tk_{\cT}(\x_{\cT}, \X_{\cT}) \right).
\end{align*}

Now, leveraging the additive structure of the kernel, we express each term via recursive constructions. Let
\[
\z_j = \tk_j(x_j, \X_j)
\]
and define the sets:
\[
\cZ_{-i} = \{ \z_j : j \in \cD \setminus \{i\} \}.
\]
The elementary symmetric polynomials (ESPs) over \(\cZ_{-i}\) are defined recursively as:
\[
e_0(\cZ_{-i}) = \onevec, \,
e_r(\cZ_{-i}) = \frac{1}{r} \sum_{s=1}^r (-1)^{s-1} e_{r-s}(\cZ_{-i}) \odot p_s(\cZ_{-i}),
\]
with the element-wise power sum being defined as:
\[
p_s(\cZ_{-i}) = \sum_{\z \in \cZ_{-i}} \z^s.
\]

Using this structure, the sum over \(\cS \ni i\) can be factorized as:
\[
\sum_{\substack{\cS \subseteq \cD \\ \cS \ni i}} \w{\cS} \cdot \tk_{\cS}(\x_{\cS}, \X_{\cS}) = \z_i \odot \sum_{q=0}^{d-1} w_{q+1} e_q(\cZ_{-i}).
\]
Define the shorthand:
\[
\pmb\ell_i := \z_i \odot \sum_{q=0}^{d-1} w_{q+1} e_q(\cZ_{-i}).
\]

\paragraph{Part 1: Mean.}
Substituting into the expression for the mean, we obtain:
\[
\bE_f(\phi_i \mid \postdata) = \pmb\ell_i^\top \alphab.
\]

\paragraph{Part 2: Variance.}
The first term in the variance is similar to $\pmb \ell_i$, and can be thus rewritten as follows using ESPs:
\[
\sum_{\substack{\cS \subseteq \cD \\ \cS \ni i}} \frac{\sigma_{|\cS|}^2}{|\cS|^2} \cdot \tk_{\cS}(\x_{\cS}, \x_{\cS}) 
= \bar{z}_i \cdot \sum_{q=0}^{d-1}\frac{\sigma_{q+1}^2}{(q+1)^2} e_{q}(\bar{\cZ}_{-i}),
\]
where $e_{q}(\bar{\cZ}_{-i})$ is the ESP for set $\bar{\cZ}_i$. Combining everything, we get the closed-form variance:
\[
\V_f(\phi_i \mid \postdata) = 
\left( \bar{z}_i \cdot \sum_{q=0}^{d} \tfrac{\sigma^2_{q+1}}{(q+1)^2}  e_{q}(\bar{\cZ}_{-i}) \right)
- \pmb\ell_i^\top \Sigma^{-1} \pmb \ell_i,
\]
as claimed.

\subsection{Proof of Theorem \ref{th:sv global}}
To prove the theorem, we need to find a closed-form formula for $m_G$, and the Shapley value will follow directly from that. In particular, $m_G(\cS) = \V_x(f_{\cS}(\x_{\cS}))$, the global Shapley value is then computed as:
\begin{align}\label{eq:sv var}
    \phi_i(v_G) = \sum_{\substack{\cS \subseteq \cD \\ \cS \ni i}} \frac{m_G(\cS)}{|\cS|} = \sum_{\substack{\cS \subseteq \cD \\ \cS \ni i}} \frac{1}{|S|} \V_x(f_{\cS}(\x_{\cS}))
\end{align}

We now need to compute $\V_X(f_S(\x))$: 

\begin{align}\label{eq:var f_s}
\V_X&(f_{\cS}(\x_{\cS})) = \V_X\left(\bigg(\sigma^2_{|\cS|} \bigodot_{j\in \cS} \tk_j(x_j,\X_j)\bigg)^\top \alphab \right)\cr
&= \sigma_{|\cS|}^4\alphab^\top \Cov \left(\bigodot_{i \in \cS}
\tk_i(x_i, \X_i)\right) \alphab \cr 
&= \sigma_{|\cS|}^4\alphab^\top \left(\bigodot_{i \in \cS} \int_{\mathcal{X}_i} \tk_i(x_i, \X_i) \tk_i(x_i,\X_i)^\top dp(x_i)\right) \alphab.
\end{align}

Replacing equation \eqref{eq:var f_s} in equation \eqref{eq:sv var}, we get
{\small
\begin{align*}
        \phi_i(v_G) = \alphab^{\!\top} \bigg(\sum_{\substack{\cS\subseteq \cD \\ S \ni i}} \frac{\sigma^4_{|\cS|}}{|\cS|} \bigodot_{j \in \cS} \int_{\mathcal{X}_j} \tk_i(x_j, \X_j) \tk_i(x_j,\X_j)^{\!\top} dp(x_j)\bigg) \alphab
\end{align*}
}
and that completes the proof.

\subsection{Proof of Theorem \ref{th:global recursion}}
We define $\M_i$ as
\[
\M_i :=
\sum_{\cS \subseteq \cD, \cS \ni i}
\frac{\sigma_{|\cS|}^4}{|\cS|}\,
\bL_{\cS}=
\sum_{k=1}^{d}
\frac{\sigma_k^4}{k}
\sum_{\substack{\cS \subseteq \cD\\|\cS|=k,\,\cS \ni i}}
\bigodot_{j\in \cS}\bL_j.
\]
In each inner sum pick out the term for $i$ and write $\cS=\{i\}\cup \cT$ with $\cT\subseteq \cD\setminus\{i\}$, $|\cT|=k-1$.  Then
$
\bigodot_{j\in \cS}\bL_j
=
\bL_i \odot\bigodot_{j\in \cT}\bL_j,
$
so that
$
\M_i
=
\sum_{k=1}^{d}
\frac{\sigma_k^4}{k}\,
\bL_i
\odot
\sum_{\substack{\cT\subseteq \cD\setminus\{i\}\\|\cT|=k-1}}
\bigodot_{j\in \cT}\bL_j.
$
Reindexing by $r=k-1$ gives
\[
\M_i
=
\bL_i \odot
\sum_{r=0}^{d-1}
\frac{\sigma_{r+1}^4}{r+1}
\underbrace{\sum_{\substack{\cT\subseteq \cD\setminus\{i\}\\|\cT|=r}}
\bigodot_{j\in \cT}\bL_j}_{e_r(\cZ_{-i})},
\]
where by definition the inner sum is precisely the order-$r$ elementary symmetric–matrix polynomial
of the set $\cZ_{-i} = \{\bL_j : j \in \cD \setminus \{i\} \}$.  Finally, by Newton’s identities, we have for each $r\ge1$
$
e_r(\cZ_{-i})
=
\frac{1}{r}
\sum_{s=1}^r
(-1)^{s-1}\,
e_{\,r-s}(\cZ_{-i})
\odot
p_s(\cZ_{-i})
\quad\text{and}\quad
e_0(\cZ_{-i})=\onevec,
$
which completes the derivation of
$
\M_i
=
\bL_i\odot
\sum_{r=0}^{d-1}
\frac{\sigma_{r+1}^4}{r+1}\,
e_r(\cZ_{-i})
$
and hence of the formula
$
\phi_i(v_G)
=
\alphab^\top
\Bigl(\M_i\Bigr)\alphab.
$

\subsection{Proof of Lemma \ref{lem:global additivity}}
Using the \mobius representation $m(\cS) := \V_X(f_{\cS}(\x_{\cS}))$, the Shapley value of feature $i$ with respect to the variance-based value function $v_G$ is:
\[
\phi_i(v_G) = \sum_{\substack{\cS \subseteq \cD \\ \cS \ni i}} \frac{m(\cS)}{|\cS|}.
\]
Summing over all features yields:
\begin{align*}
\sum_{i=1}^d \phi_i(v_G) &= \sum_{\substack{\cS \subseteq \cD \\ \cS \neq \varnothing}} \left( \sum_{i \in \cS} \frac{1}{|\cS|} \right) m(\cS) \cr 
&= \sum_{\substack{\cS \subseteq \cD \\ \cS \neq \varnothing}} m(\cS) \cr 
&= \sum_{\substack{\cS \subseteq \cD \\ \cS \neq \varnothing}} \V_X(f_{\cS}(\x_{\cS})) \cr
&= \V_X(f(\x)) - \V_X(f_{\varnothing}),
\end{align*}
where the third equality uses the fact that for any non-empty set $\cS$, $\sum_{\cS \ni i} \frac{1}{|\cS|} = 1$, and the final equality removes the contribution of the constant (zero-order) term $f_{\varnothing}$, which has zero variance.

Since the total variance of the output is:
\[
\V_X(y) = \V_X(f(\x)) + \sigma_n^2,
\]
it follows that:
\[
\V_X(y) - \sigma_n^2 = \V_X(f(\x)) = \sum_{i=1}^{d} \phi_i(v_G).
\]

\section{Newton’s Identities}\label{apx:newton identities}
Newton’s identities establish a fundamental relationship between two important families of symmetric polynomials: the \emph{elementary symmetric polynomials (ESPs)} and the power sum polynomials. Let \(\mathcal{Z}_d = \{z_1, z_2, \ldots, z_d\}\) be a set of variables. The ESPs of degree \(q\), denoted \(e_q\), is defined as the sum over all products of \(q\) distinct variables from \(\mathcal{Z}_d\):
\[
e_q = \sum_{1 \leq i_1 < i_2 < \cdots < i_q \leq d} z_{i_1} z_{i_2} \cdots z_{i_q},
\]
with the conventions \(e_0 = 1\) and \(e_q = 0\) for \(q > d\). In contrast, the power sum polynomial of degree \(q\), denoted \(p_q\), is defined as the sum of the \(q\)-th powers of the variables:
\[
p_q = z_1^q + z_2^q + \cdots + z_d^q.
\]

Newton’s identities provide a recursive method to express the elementary symmetric polynomials in terms of the power sums. Specifically, for each integer \(q \geq 1\), the identity takes the form
\[
q e_q = \sum_{j=1}^q (-1)^{j-1} e_{q-j} p_j.
\]
Equivalently, one can isolate \(e_q\) as
\[
e_q = \frac{1}{q} \sum_{j=1}^q (-1)^{j-1} e_{q-j} p_j.
\]
These identities hold for all \(q \leq d\), and because each \(e_q\) depends only on the previous elementary polynomials and the power sums, the recursion provides an efficient means to compute one family from the other.

To illustrate, consider the case \(d = 3\), with variables \(z_1, z_2, z_3\). The first few elementary symmetric polynomials are:
\[
e_1 = z_1 + z_2 + z_3, \quad
e_2 = z_1 z_2 + z_1 z_3 + z_2 z_3, \quad
e_3 = z_1 z_2 z_3,
\]
and the corresponding power sums are
\[
p_1 = z_1 + z_2 + z_3, \quad
p_2 = z_1^2 + z_2^2 + z_3^2, \quad
p_3 = z_1^3 + z_2^3 + z_3^3.
\]
Applying Newton’s identities, we compute:
\[
e_1 = p_1,
\]
\[
2e_2 = e_1 p_1 - p_2 = p_1^2 - p_2 \quad \Rightarrow \quad e_2 = \frac{1}{2}(p_1^2 - p_2),
\]
\[
3e_3 = e_2 p_1 - e_1 p_2 + p_3.
\]
Substituting the earlier results, this gives a full expression for \(e_3\) in terms of \(p_1, p_2,\) and \(p_3\).

\section{Efficient and Stable Algorithm for the Mean and Covariance of SSVs}\label{apx:ssv-efficient}

\subsection{Efficient Algorithm of SSVs: Mean and Covariance}
The stochastic Shapley values \(\pmb\phi(\nu_{\x}) = [\phi_1, \dots, \phi_d]^\top\) follow a multivariate Gaussian distribution with mean vector and covariance matrix defined in Proposition~\ref{prop:gp sv}. While these quantities involve exponential sums over subsets \(\cS \subseteq \cD\), the structure of FANOVA GPs allows for a recursive formulation that enables efficient computation.

\paragraph{Mean.}  
As shown in Theorem~\ref{th:mean-var recursion}, the mean of the SSV for feature \(i\) can be written in terms of the elementary symmetric polynomials (ESPs) over the kernel terms:
\[
\pmb\ell_i := \z_i \odot \sum_{q=0}^{d-1} w_{q+1} e_q(\cZ_{-i}),
\quad \text{where } \z_j := \tilde{k}_j(x_j, \X_j).
\]
Then, the mean is computed as
\[
\xi_{\phi_i} = \pmb\ell_i^\top \alphab.
\]

\paragraph{Covariance.}
The covariance between SSVs \(\phi_i\) and \(\phi_j\) consists of two terms:
\[
\Cov(\phi_i, \phi_j)
= \sum_{\substack{\cS \subseteq \cD \\ \{i,j\} \subseteq \cS}} \frac{\sigma_{|\cS|}^2}{|\cS|^2} \tk_{\cS}(\x_{\cS}, \x_{\cS})
- \pmb\ell_i^\top \Sigma^{-1} \pmb\ell_j.
\]
The second term is efficiently computed once \(\pmb\ell_i\) and \(\pmb\ell_j\) are known. To compute the first term efficiently, observe that we can factor the interaction term \(\tk_{\cS}(\x_{\cS}, \x_{\cS})\) as:
\[
\tk_{\cS}(\x_{\cS}, \x_{\cS}) = \tk_i(x_i, x_i)\tk_j(x_j, x_j) \cdot \tk_{\cS \setminus \{i,j\}}(\x_{\cS \setminus \{i,j\}}, \x_{\cS \setminus \{i,j\}}).
\]
Let \(\bar{z}_j := \tk_j(x_j, x_j)\), and define \(\bar{\cZ}_{-\{i,j\}} := \{\, \bar{z}_k : k \in \cD \setminus \{i,j\} \}\). Then the sum over subsets \(\cS \ni \{i,j\}\) can be expressed recursively via ESPs:
\[
\sum_{\substack{\cS \subseteq \cD \\ \cS \ni \{i,j\}}} \frac{\sigma_{|\cS|}^2}{|\cS|^2} \tk_{\cS}(\x_{\cS}, \x_{\cS})
= \bar{z}_i \bar{z}_j \cdot \sum_{q=0}^{d-2} \frac{\sigma_{q+2}^2}{(q+2)^2} e_q(\bar{\cZ}_{-i,j}),
\]
where the ESPs are defined recursively:
\begin{align*}
&e_0(\bar{\cZ}_{-\{i,j\}}) = 1, \\
&e_r(\bar{\cZ}_{-\{i,j\}}) = \frac{1}{r} \sum_{s=1}^{r} (-1)^{s-1} e_{r-s}(\bar{\cZ}_{-i,j}) \cdot p_s(\bar{\cZ}_{-\{i,j\}}),
\end{align*}
with power sums \(p_s(\bar{\cZ}_{-\{i,j\}}) = \sum_{\bar{z} \in \bar{\cZ}_{-\{i,j\}}} \bar{z}^s\).

The combination of the ESP-based recursion and the precomputed vectors \(\pmb\ell_i\) enables efficient computation of all entries in the mean and covariance matrix. Algorithm \ref{alg:shapley-mean-cov} summarizes the overall algorithm for computing the mean and covariance of stochastic Shapley values for explaining an instance.

\begin{algorithm}[H]
\caption{Efficient computation of mean and covariance of stochastic Shapley values}
\label{alg:shapley-mean-cov}
\begin{algorithmic}[1]
\REQUIRE Data $\X$, test input $\x$, kernel components $\tk_j$, GP posterior mean $\alphab$ and $\Sigma$, scale parameters $\{\sigma_q\}_{q=0}^d$
\ENSURE Mean vector \(\pmb \xi\) and covariance matrix \(\mathbf{K}_\phi\)

\FOR{$i = 1$ to $d$}
    \STATE Compute $\z_j := \tk_j(x_j, \X_j)$ for $j \ne i$
    \STATE Compute ESPs $e_q(\cZ_{-i})$ over $\cZ_{-i} = \{\z_j : j \ne i\}$ using Newton's identities
    \STATE Compute $\pmb\ell_i := \z_i \odot \sum_{q=0}^{d-1} w_{q+1} e_q(\cZ_{-i})$
    \STATE Compute mean $\xi_{\phi_i} := \pmb\ell_i^\top \alphab$
\ENDFOR

\FOR{$i = 1$ to $d$}
    \FOR{$j = i$ to $d$}
        \STATE Compute $\bar{z}_k := \tk_k(x_k, x_k)$ for $k \notin \{i,j\}$
        \STATE Compute ESPs $e_q(\bar{\cZ}_{-i,j})$ over $\bar{\cZ}_{-i,j} = \{\bar{z}_k : k \notin \{i,j\}\}$
        \STATE Compute first term:
        \[
        V_{i,j}^{(1)} := \bar{z}_i \bar{z}_j \cdot \sum_{q=0}^{d-2} \tfrac{\sigma_{q+2}^2}{(q+2)^2} e_q(\bar{\cZ}_{-i,j})
        \]
        \STATE Compute second term:
        \[
        V_{i,j}^{(2)} := \pmb\ell_i^\top \Sigma^{-1} \pmb\ell_j
        \]
        \STATE Set $[\mathbf{K}_\phi]_{i,j} = V_{i,j}^{(1)} - V_{i,j}^{(2)}$
        \STATE Set $[\mathbf{K}_\phi]_{j,i} = [\mathbf{K}_\phi]_{i,j}$ \COMMENT{Symmetry}
    \ENDFOR
\ENDFOR

\STATE \RETURN $\pmb\xi, \mathbf{K}_\phi$
\end{algorithmic}
\end{algorithm}

\subsection{Numerically Stable Computation of ESPs via Characteristic Polynomials}\label{apx:stable esp}

The recursive computation of ESPs using Newton's identities, though elegant, can be numerically unstable, particularly when the values involved span several orders of magnitude or when many features are involved. This instability arises from alternating signs and divisions by increasing integers, which may lead to cancellation errors and loss of precision.

To mitigate this, we adopt a numerically stable alternative based on polynomial expansions. Specifically, given a set of values $\cZ = \{z_1, z_2, \ldots, z_n\}$, the ESPs can be obtained as the coefficients in the expansion of the following generating polynomial:
\begin{equation}
    P(t) = \prod_{j=1}^n (1 + z_j t) = \sum_{q=0}^n e_q(\cZ) t^q.
\end{equation}
This expression defines a univariate polynomial whose coefficient of $t^q$ is exactly the ESP of order $q$ over $\cZ$. That is, the $q$-th ESP $e_q(\cZ)$ is the sum over all products of $q$ distinct elements in $\cZ$, matching its combinatorial definition.

The computation proceeds iteratively, rather than recursively, and avoids divisions or subtraction of nearly equal quantities. The algorithm below computes the ESPs using a simple update rule.

\begin{algorithm}[H]
\caption{Stable computation of ESPs via polynomial expansion}
\label{alg:stable-esp}
\begin{algorithmic}[1]
\REQUIRE A list of values $\cZ = \{z_1, z_2, \dots, z_n\}$
\ENSURE ESPs $[e_0, e_1, \dots, e_n]$
\STATE Initialize $e[0] \gets 1$, and $e[q] \gets 0$ for $q = 1$ to $n$
\FOR{each $z$ in $\cZ$}
    \FOR{$q$ from $n$ down to $1$}
        \STATE $e[q] \gets e[q] + z \cdot e[q-1]$
    \ENDFOR
\ENDFOR
\RETURN $e[0:n+1]$
\end{algorithmic}
\end{algorithm}

This scheme ensures that each ESP of order $q$ is constructed incrementally using lower-order terms, maintaining numerical robustness. In our implementation of Shapley value computation (see Algorithm~\ref{alg:shapley-mean-cov}), we use this stable ESP routine to replace the Newton-based formulation when evaluating the recursive expressions in equation~\eqref{eq:l recursion} and the global variant.

\section{Experiments}\label{apx:experiments}

\subsection{Experimental Setup}\label{apx:methods setup}

\paragraph{FANOVA GP Training Setup}\label{apx:AGP setup}
Missing features in the input data are imputed using the mean strategy. The FANOVA GP model is configured with a maximum interaction depth of five for both synthesized and real experiments.

For \textit{regression}, labels are standardized using a standard scaler, and the Gaussian likelihood is employed for the FANOVA GP. An RBF kernel serves as the base kernel and is extended to an orthogonal RBF kernel for experiments involving functional decomposition. Normalizing flow is applied to the features, transforming them into a normal distribution with a mean of zero and a standard deviation of one. This transformation facilitates the derivation of an analytical formula for the orthogonal RBF kernel. Additionally, sparse Gaussian processes are utilized for training, with non-trainable inducing points determined through a clustering algorithm, similar to the approach described in \cite{AGP_ortho}. For datasets \texttt{pymadyn32nm} and \texttt{keggdirect}, 800 inducing points are used, while 200 inducing points are fixed for other datasets.

For \textit{binary classification}, the FANOVA GP is trained using a Bernoulli likelihood. A \textit{Stochastic Variational Gaussian Process (SVGP)} is employed for mini-batch training, with a mini-batch size of 512 or the dataset size (whichever is smaller). The model is trained using the Adam optimizer with a learning rate of 0.01, and the training process spans 500 epochs. The inducing points are initialized using k-means clustering and remain fixed throughout the training process.

For classification tasks, we apply our explainability method to the latent FANOVA GP function before the Bernoulli likelihood transformation. This ensures that the explanations reflect the underlying continuous function, on which our method can be applied.

\paragraph{Explainable Methods Setup}\label{apx:exp setup}
For generating explanations with SHAP, BiSHAP, Sampling SHAP, LIME, and MAPLE, we use 1024 samples to estimate feature importance values. As suggested in SHAP, we apply SHAP clustering to select a representative background dataset and use 100 samples as the background for computing value functions. This same set of background samples is used consistently across all methods to ensure a fair comparison. MAPLE adapts its local models using the provided background data, ensuring comparable conditions across methods.

\paragraph{Feature Selection Methods Setup}\label{apx:fs setup}
For HSICLasso, we use the RBF kernel for features and labels in regression tasks, and the categorical kernel for classification tasks, while keeping the remaining parameters at their default values as provided in the Python library. For Lasso, we tune the regularization parameter using 5-fold cross-validation. For the Tree Ensemble method, we use the \texttt{ExtraTreeClassifier} from the \texttt{scikit-learn} library with 500 estimators, and feature importance is determined using a permutation test. For the first-order Sobol index, we use the implementation by \citet{AGP_ortho}, but only include the importance of individual features, not their interactions (as the typical choice for first-order Sobol index).

\paragraph{Training Random Forest with Grid Search}\label{apx:rf setup}

In this study, a Random Forest model is trained for either classification or regression tasks, depending on the problem. The training process involves handling missing values by imputing the mean and optimizing the model's performance through hyperparameter tuning. The parameters considered for tuning include the number of trees in the forest (500 and 1000) and the maximum depth of the trees (None, 10, and 20), which influence the model's complexity and performance.

To ensure robust evaluation, a grid search with cross-validation is performed on 90\% of the data to identify the best combination of hyperparameters. The model's performance is then evaluated on the remaining 10\% test data using accuracy for classification tasks or mean absolute percentage error for regression tasks.

\subsection{Synthetic Experiments}\label{apx:synthesized experiment}
\subsubsection{Dataset Generation}\label{apx:synthetic dataset}
In this study, we generate four synthetic datasets to evaluate and compare feature selection and explanation methods. The predictors \(X\) consist of 1000 samples with 20 independent features drawn from a Gaussian distribution. For each dataset, the response variable \(Y\) is generated using a unique function to capture different types of interactions and non-linear relationships between features. The datasets are described as follows:

\emph{Synthetic Dataset 1} The response variable \(Y\) is modeled as:
\[
Y = X_1^2 - 0.5 X_2^2 + \sin(2\pi X_1),
\]
where only the first two features (\(X_1\) and \(X_2\)) are relevant. The dataset includes polynomial effects and nonlinear interactions (\(\sin(2\pi X_1)\)) to evaluate the detection of complex relationships. The expected average rank of the most important features is ideally 1.5.

\emph{Synthetic Dataset 2} The response variable \(Y\) is expressed as:
\begin{align*}
Y &= \exp(X_1) \tanh(X_2 X_3) + \exp(-|X_4|) \tanh(X_1 X_2) \cr 
&\qquad + \exp(X_1 X_2) \sin(X_3 X_4),
\end{align*}
where the first four features (\(X_1\), \(X_2\), \(X_3\), \(X_4\)) significantly influence the outcome. This dataset introduces nested exponential and hyperbolic functions to evaluate feature selection under complex conditions. The expected average rank for the most important features is ideally 2.5.

\emph{Synthetic Dataset 3} The response variable \(Y\) is modeled as:
\begin{align*}
Y &= \sin(X_1) \exp(X_2) + \cos(X_3 X_4) \tanh(X_1 X_2 \pi) \cr 
&\qquad + \exp\left(-(X_1^2 + X_2^2)\right) \sin((X_3 + X_4)\pi),
\end{align*}
where the first four features (\(X_1\), \(X_2\), \(X_3\), \(X_4\)) contribute to the outcome. The dataset captures a mix of trigonometric, exponential, and interaction effects. The expected average rank of the most important features is ideally 2.5.

\emph{Synthetic Dataset 4} The response variable \(Y\) is expressed as:
\[
Y = \exp\left(\sum_{i=1}^3 X_i^2 - 4\right),
\]
making the first three features (\(X_1\), \(X_2\), \(X_3\)) particularly significant. This dataset emphasizes the importance of features contributing exponentially to the outcome. The expected average rank for the most important features is ideally 2.

\subsubsection{Evaluation on Synthetic Datasets Using average Rank}\label{apx:synthetic evaluation}
For each synthesized dataset, the most important features are predefined based on the function used to generate the response variable \(Y\). The evaluation process is carried out as follows:

\begin{enumerate}
    \item \emph{Sampling:} For each dataset, 1000 samples are randomly selected, and explanations are generated for these samples using various local explainer methods, as well as the proposed algorithm here.
    
    \item \emph{Ranking Features:} For each sample, the features are ranked based on their absolute importance scores. Features with higher importance scores are assigned lower ranks (e.g., rank 1 for the most important feature).
    
    \item \emph{Average Rank of Important Features:} For each sample, the ranks of the known most important features are extracted, and their average is computed. This mean gives a single average rank for the sample.
    
    \item \emph{Aggregate Statistics:} Across the 500 samples, 500 average ranks are obtained for each method. The distribution of these average ranks is then visualized using a box plot. The box plot provides insights into the consistency and effectiveness of the explainer in identifying the most important features.
\end{enumerate}

For instance, in the first synthetic dataset, the first two features (\(X_1\) and \(X_2\)) are predefined as the most important. After generating 500 samples, the local explainers assign importance scores to all features for each sample. These scores are ranked, and the average rank of \(X_1\) and \(X_2\) is computed for each sample. This results in 500 average ranks, which are aggregated into a box plot to evaluate the explainers' performance. This process is repeated for all synthetic datasets.

We adopt a similar strategy to compare feature selection methods on the synthetic dataset. However, the feature selection process is repeated 30 times, for each of which we generate a whole data set and the compute the average rank of the features accordingly. The results are also visualized using a box plot to illustrate the distribution and variability of the average ranks across the repetitions.


\subsection{Experiments on Real Datasets}

\subsubsection{Explanation}
We assess \fgpxl for local explanations in real-world data sets by analyzing the impact of removing the most important features identified by local Shapley values. The additive structure of FANOVA GP allows us to compute the prediction of a trained model with only a selected subset of features (when the FANOVA GP is trained on \textit{all} features). To compute the prediction of an FANOVA GP for a subset of features, we adapt the additive kernel to only include features and interactions involving the selected features. Let $S \subseteq D$ represent the subset of features of interest, and denote the input restricted to these features by $\x_S$. The additive kernel for the subset $S$ is defined as:
\begin{align}\label{eq:additive kernel subset}
    k^{add_S}(\x_{\cS}, \x'_{\cS}) = \sum_{q=0}^{|\cS|} k^{add_{q,\cS}}(\x_{\cS}, \x'_{\cS}),
\end{align}
where $k^{add_{q,\cS}}(\x_{\cS}, \x'_{\cS})$ is the $q$-th order additive kernel restricted to feature subset $\cS$, computed as:
\begin{align}\label{eq:additive kernel order subset}
    k^{add_{q,\cS}}(\x_{\cS}, \x'_{\cS}) = \sigma_q^2 \sum_{i_1 \leq i_2 \leq \cdots \leq i_q,\, i_l \in \cS} \left[ \prod_{l=1}^q k_{i_l}(\x_{\cS}, \x'_{\cS}) \right],
\end{align}
where $\sigma_q^2$ is the scale assigned to interactions of order $q$ involving only the features in $\cS$. The subset kernel $k^{add_{\cS}}(\x_{\cS}, \x'_{\cS})$ captures the additive structure within the selected subset of features, ensuring that interactions outside $\cS$ are ignored. The prediction for sample $\x_{\cS}$ using the restricted additive kernel is given by:
\begin{align}\label{eq:gp subset prediction}
    f_{\cS}(\x) = k^{add_{\cS}}(\x_{\cS}, \X_{\cS})^\top \alphab,
\end{align}
where $k^{add_{\cS}}(\x_{\cS}, \X_{\cS})$ is the additive kernel matrix computed over the training data $X_{\cS}$ restricted to features in $S$, $k^{add_S}(\x_{\cS}, X_{\cS})$ is the additive kernel vector between the sample $\x_{\cS}$ and the training data $\X_{\cS}$, and all the other parameters are used from the trained FANOVA GP on all the datasets. This formulation allows a trained model to provide predictions based on only the selected subset of features ${\cS}$, making it suitable for scenarios where we want to verify the effect of removing features (see Experiments for more information).

\begin{figure*}[t]
    \centering
    \includegraphics[width=0.95\linewidth]{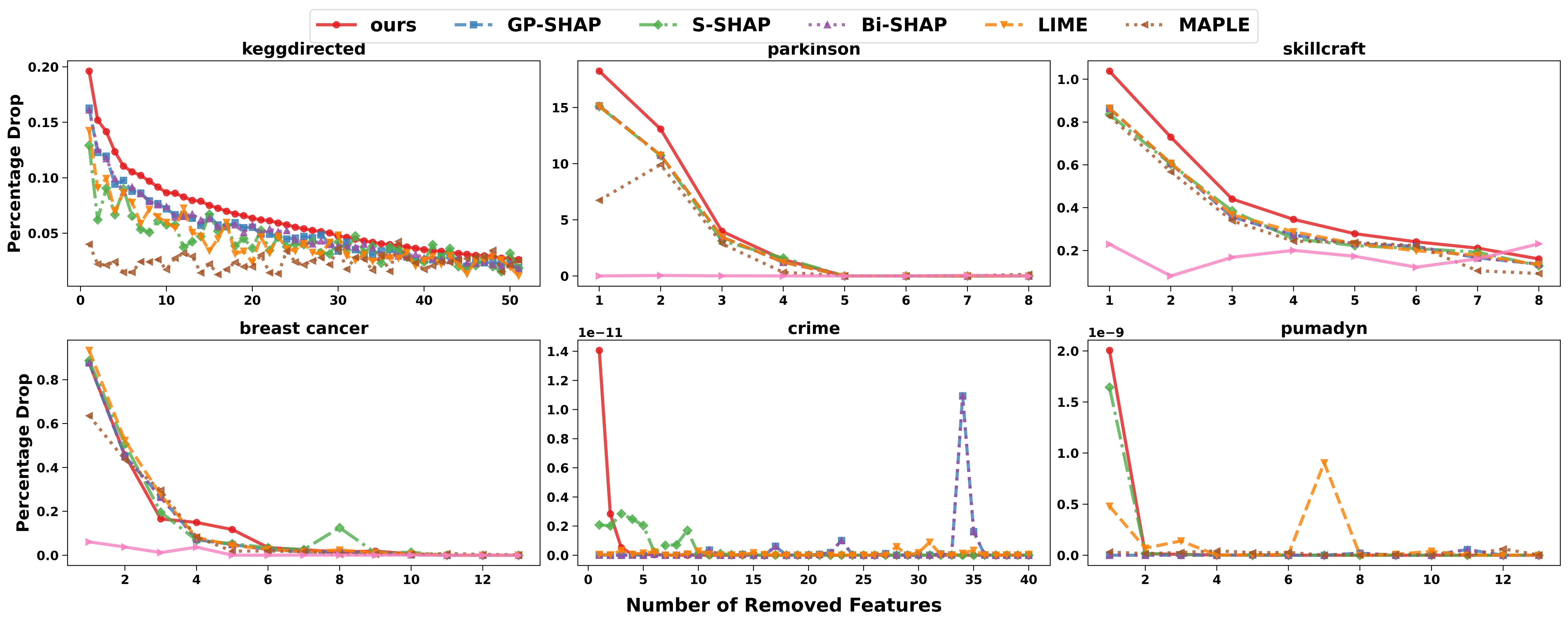}
    \caption{The percentage drop in the prediction by removing features.}
    \label{fig:exp real feature removal}
\end{figure*}

\begin{table*}[!ht]
    \centering
    \caption{The performance of random forest on nine data sets using top 20\% features idenfied by different feature selectors.}
    \footnotesize
    \label{tab:fs performance}
    \resizebox{\textwidth}{!}{
    \begin{tabular}{l| ccc | cccccc}
        ~ & \multicolumn{3}{c|}{classification (accuracy $\uparrow$)}  & \multicolumn{6}{c}{ regression (MAPE $\downarrow$)} \\ 
        Feature Selector & sonar & nomao  & wisconsin & crime & breast cancer & pumadyn & skillcraft & parkinson & keggdirected \\ \hline
        \fgpxg & \textbf{0.95} & \textbf{0.95} & \textbf{0.96} & \textbf{0.18} & \textbf{0.29} & \textbf{1.01} & \textbf{0.20} & \textbf{0.01} & 0.13 \\ 
        Sobol & 0.76 & 0.86 & 0.91 & \textbf{0.18} & 0.60 & 1.04 & 0.25 & 0.95 & \textbf{0.12} \\ 
        HSICLasso & 0.81 & 0.92 & 0.89 & \textbf{0.18} & 0.47 & 1.08 & \textbf{0.20} & 0.90 & \textbf{0.12} \\ 
        MI & 0.86 & 0.93 & 0.95 & \textbf{0.18} & 0.51 & 1.04 & 0.21 & \textbf{0.01} & 0.15 \\ lasso & 0.90 & 0.92 & 0.95 & 0.20 & 0.52 & 1.11 & 0.21 & 0.87 & 0.14 \\ 
        F-ANOVA & 0.81 & 0.92 & 0.95 & \textbf{0.18} & 0.55 & 1.04 & 0.21 & 0.87 & 0.13 \\
        TE & 0.90 & 0.94 & \textbf{0.96} & \textbf{0.18} & 0.37 & \textbf{1.01} & \textbf{0.20} & \textbf{0.01} & 0.13 \\
        \hline
        \end{tabular}}
\end{table*}

For this experiment, we calculate the percentage drop in predictions after feature removal. A greater drop indicates the removal of more critical features. The evaluation of feature removal methods is based on measuring the impact of removing each feature on the model's predictions. For a given sample, the initial prediction before any features are removed is denoted as \(x_0\). The prediction after removing the \(t\)-th feature is represented as \(x_t\). The effect of feature removal is quantified by computing the normalized drop in prediction between consecutive feature removals. Specifically, the normalized drop is defined as:

\[
d_t = \frac{|x_{t+1} - x_t|}{|x_0| + \epsilon},
\]

where \(d_t\) represents the normalized drop after removing the \((t+1)\)-th feature, and \(\epsilon\) is a small constant added to \(x_0\) to ensure numerical stability and prevent division by zero. This formulation captures the relative change in prediction caused by feature removal, scaled by the magnitude of the original prediction. By normalizing the differences in this manner, the measure becomes comparable across samples, even when their baseline predictions vary significantly.

For each method, the normalized drop values \(d_t\) are computed across all samples for each feature removal step, and their mean is calculated to summarize the method's overall behavior. The mean normalized drop, \(\bar{d}_t\), provides an average measure of the sensitivity of the model's predictions to the removal of each feature. This is computed as:

\[
\bar{d}_t = \frac{1}{N} \sum_{i=1}^{N} d_{t,i},
\]

where \(N\) is the total number of samples, and \(d_{t,i}\) is the normalized drop for the \(i\)-th sample at step \(t\). By examining the trend of \(\bar{d}_t\) across feature removal steps, we can evaluate how effectively each method prioritizes important features. Methods that identify the most critical features exhibit larger drops in prediction performance during the early stages of feature removal, as the most impactful features are removed first. This framework provides a consistent and interpretable way to compare feature removal methods across datasets.

\begin{figure*}[t]
    \centering
    \includegraphics[width=0.97\linewidth]{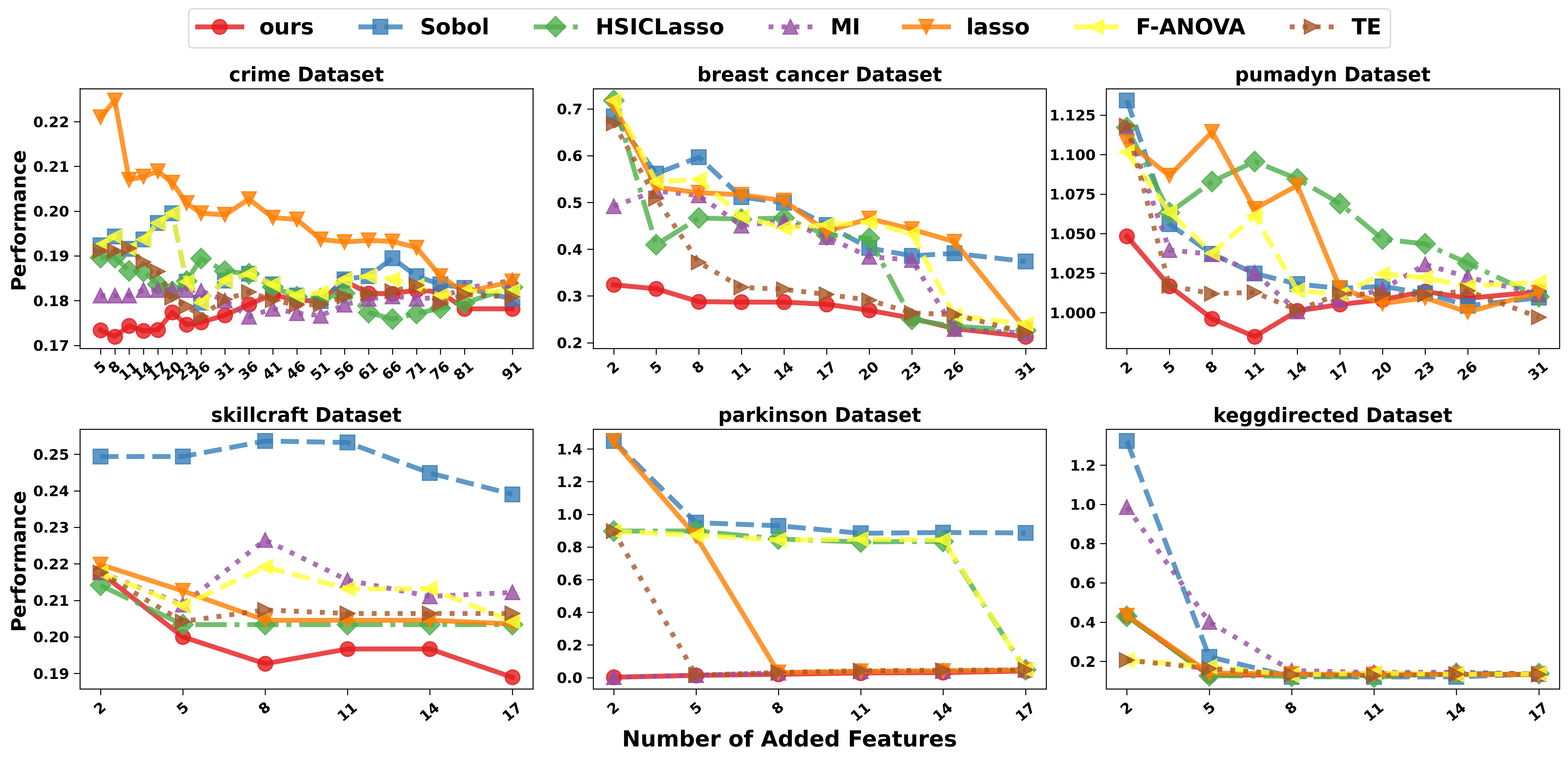}
    \caption{The performance of random forest (i.e., MAPE) as a function of features added based on their importance by different feature selectors.}
    \label{fig:fs real ds}
\end{figure*}

\begin{figure*}[t]
    \centering
    \includegraphics[width=1\textwidth]{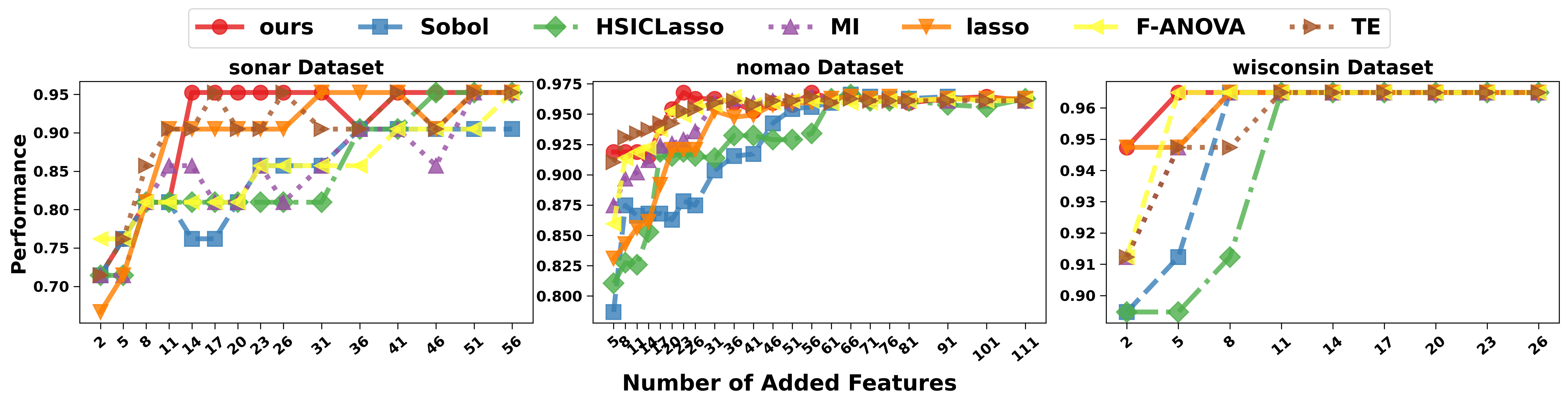}
    \caption{The performance of random forest (i.e., accuracy) by adding more features based on their importance by different feature selectors.}
    \label{fig:fs real classification}
\end{figure*}

A FANOVA GP is trained on each dataset, and 50 instances are randomly selected for explanation. Explanations are generated using different methods, based on which the most important features are removed. The overall impact of removing a feature is summarized by measuring the average percentage drop in predictions across the 50 selected instances. Methods that correctly prioritize the most influential features are expected to show a larger prediction drop early in the process, reflecting the features' significance to the model's decisions. Due to errors encountered on the majority of datasets, we excluded the U-SHAP library from our comparisons.

Figure~\ref{fig:exp real feature removal} illustrates the average prediction drop after removing each feature. \fgpxl consistently demonstrates strong performance across all datasets, reliably identifying the most impactful features. These results emphasize the superiority of \fgpxl in accurately capturing feature importance compared to other methods.

\begin{figure*}[!h]
    \centering
    \includegraphics[width=0.9\linewidth]{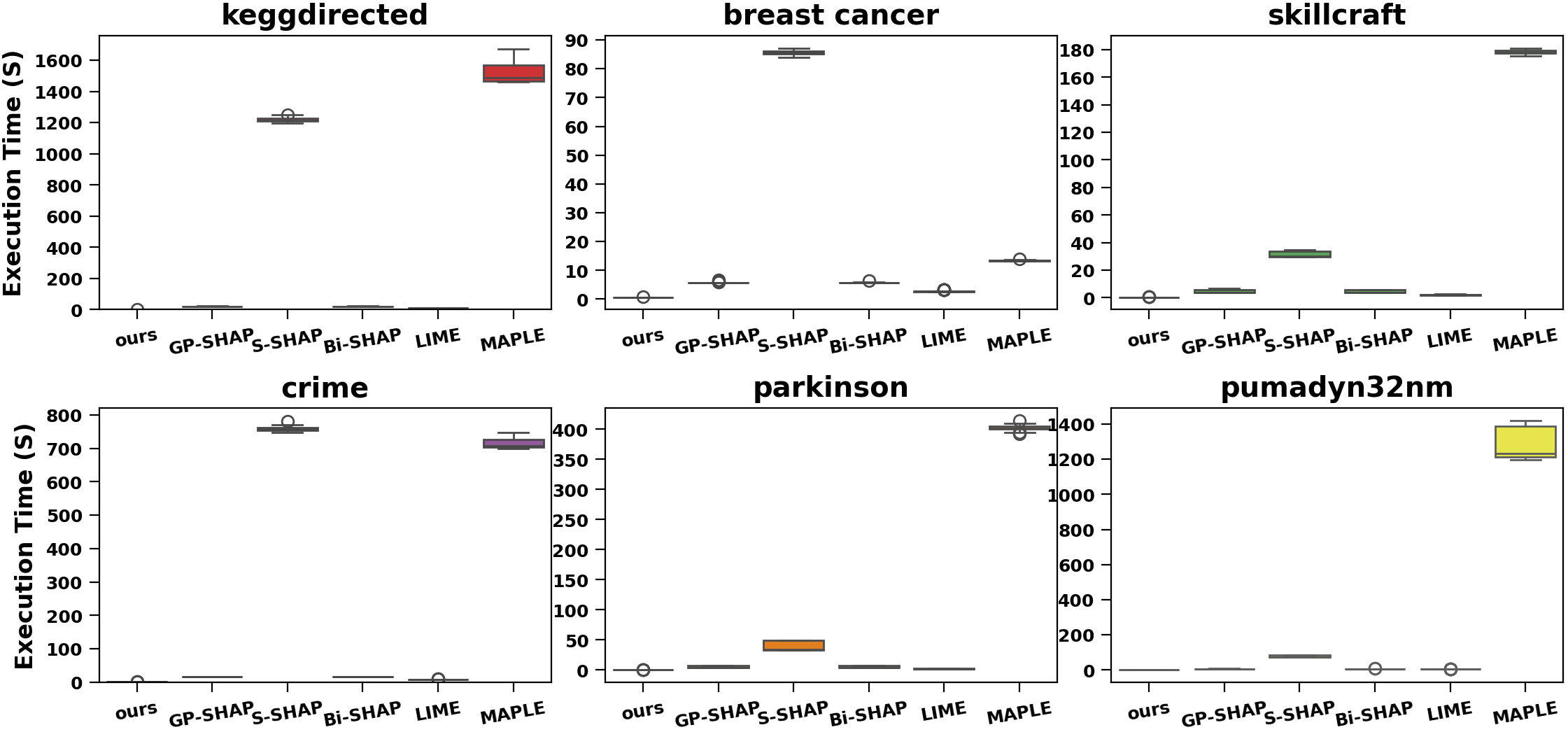}
    \caption{The time comparison of all explainable methods for generating explanations for 500 instances.}
    \label{fig:exp real time full}
\end{figure*}

\subsubsection{Feature Selection}
We evaluate \fgpxg for feature selection using real-world datasets. Each feature selector is applied to identify the importance of features in the datasets. We then select the top 20\% of the most important features identified by each method and train a random forest model using only these selected features. The rationale behind this approach is that a better feature selector identifies more relevant features, leading to improved performance of the trained random forest model (details on training and fine-tuning the random forest are discussed earlier in the section).

For classification tasks, performance is evaluated using accuracy, while for regression tasks, it is measured using the mean absolute percentage error (MAPE). Table \ref{tab:fs performance} presents the results across all nine datasets. The results demonstrate that \fgpxg consistently delivers better performance in identifying the most relevant features, excelling in both classification and regression tasks. This highlights its effectiveness as a reliable feature selection method for real-world applications.

Continuing our experiment for feature selection on real data sets, we incrementally add features based on their importance rankings and evaluate the performance of the selected features by training and fine-tuning a random forest on these subsets. We expect the most relevant features to be added early in the process, leading to an initial improvement in the random forest’s performance, which then plateaus as less relevant features are included. Figures~\ref{fig:fs real ds} and \ref{fig:fs real classification} show the performance, measured as MAPE and accuracy, as a function of the number of features added for the regression datasets. The figure demonstrates that \fgpxg consistently outperforms other feature selection methods, effectively identifying and ranking the most important features. This is evident from the significant improvement in random forest performance at the beginning of the process, observed as a lower MAPE or higher accuracy. Notably, \fgpxg performs particularly well compared to the Sobol index and surpasses other methods by accurately identifying the most relevant features in the datasets.

\subsubsection{Time Comparison}\label{apx:time}
In our experiments, we compared various methods based on their execution time, omitting some less competitive methods to provide a clearer comparison among the top-performing approaches. Figure \ref{fig:exp real time full} presents boxplots of the execution time for 500 instances across all methods. According to this plot, U-SHAP, S-SHAP, and MAPLE require significantly more time, on average, to generate explanations for a single instance.

On the other hand, LIME and GP-SHAP are notably faster. However, \fgpxl demonstrates a compelling balance of speed and performance, outperforming all other methods in terms of execution time, making it the most efficient choice for generating explanations.




\section*{Frequently Asked Questions (FAQ)}

\paragraph{Q1: Where is the implementation of your method available?}  
The full implementation of our method, including code to reproduce all experiments and figures in the paper, is publicly available at the following repository:  
\begin{center}
\url{https://github.com/Majeed7/orthogonal-additive-gaussian-processes-main}
\end{center}

\paragraph{Q2: How does your method differ from PKeX-Shapley?}  
Both our method and PKeX-Shapley use Newton's identities to compute the Shapley values in exact closed form by leveraging symmetric polynomial representations. However, there are key differences in scope and capability.

PKeX-Shapley is tailored specifically for kernel methods with product kernels. It utilizes the value function associated with functional decompositions to compute Shapley values. While this formulation enables exact computation for the mean of Shapley values, it does not extend naturally to higher-order moments. In particular, when applied to GPs, computing the variance or the covariance structure of Shapley values under product kernels is nontrivial since the orthogonality is not necessarily held in product kernels. This limitation makes it unsuitable for applications requiring uncertainty quantification or global explanations based on variance decomposition.

In contrast, our method is built on top of FANOVA GPs, where the underlying kernel has a functional ANOVA decomposition. This decomposition enables us to define a more expressive value function that captures the marginal contribution of each feature subset. However, computing Shapley values directly from this value function is intractable due to its complexity.

To address this, we introduce a (stochastic) \mobius representation of the Shapley value, which allows us to compute exact means and variances of Shapley values using recursive and efficient algorithms. This approach not only supports instance-wise explanations but also enables global explanations based on variance decomposition---something PKeX-Shapley cannot provide. Therefore, while both methods aim for exact computation, our method generalizes to richer structural insights and uncertainty quantification.

%% file: main.bbl
\begin{thebibliography}{27}
\providecommand{\natexlab}[1]{#1}
\providecommand{\url}[1]{\texttt{#1}}
\expandafter\ifx\csname urlstyle\endcsname\relax
  \providecommand{\doi}[1]{doi: #1}\else
  \providecommand{\doi}{doi: \begingroup \urlstyle{rm}\Url}\fi

\bibitem[Lundberg and Lee(2017)]{shap}
Scott~M Lundberg and Su-In Lee.
\newblock A unified approach to interpreting model predictions.
\newblock \emph{Advances in neural information processing systems}, 30, 2017.

\bibitem[Ribeiro et~al.(2016)Ribeiro, Singh, and Guestrin]{lime}
Marco~Tulio Ribeiro, Sameer Singh, and Carlos Guestrin.
\newblock " why should i trust you?" explaining the predictions of any classifier.
\newblock In \emph{Proceedings of the 22nd ACM SIGKDD international conference on knowledge discovery and data mining}, pages 1135--1144, 2016.

\bibitem[Durrande et~al.(2011)Durrande, Ginsbourger, Roustanta, and Carraro]{orthogonal_kernel}
N~Durrande, D~Ginsbourger, O~Roustanta, and L~Carraro.
\newblock Reproducing kernels for spaces of zero mean functions. application to sensitivity analysis.
\newblock \emph{stat}, 1050:\penalty0 17, 2011.

\bibitem[Hooker(2004)]{FANOVA}
Giles Hooker.
\newblock Generalized functional anova diagnostics for high-dimensional functions of dependent variables.
\newblock \emph{Journal of Computational and Graphical Statistics}, 13\penalty0 (3):\penalty0 755--770, 2004.

\bibitem[Sobol(2001)]{sobol}
Ilya~M Sobol.
\newblock Global sensitivity indices for nonlinear mathematical models and their monte carlo estimates.
\newblock \emph{Mathematics and Computers in Simulation}, 55\penalty0 (1-3):\penalty0 271--280, 2001.

\bibitem[Lu et~al.(2022)Lu, Boukouvalas, and Hensman]{AGP_ortho}
Xiaoyu Lu, Alexis Boukouvalas, and James Hensman.
\newblock Orthogonal additive gaussian processes.
\newblock In \emph{Proceedings of the 39th International Conference on Machine Learning}, pages 11956--11968, 2022.

\bibitem[Owen(2014)]{sobol_shapley}
Art~B Owen.
\newblock Sobol'indices and shapley value.
\newblock \emph{SIAM/ASA Journal on Uncertainty Quantification}, 2\penalty0 (1):\penalty0 245--251, 2014.

\bibitem[Chau et~al.(2023)Chau, Muandet, and Sejdinovic]{gp-shap}
Siu~Lun Chau, Krikamol Muandet, and Dino Sejdinovic.
\newblock Explaining the uncertain: Stochastic shapley values for gaussian process models.
\newblock \emph{Advances in Neural Information Processing Systems}, 36:\penalty0 50769--50795, 2023.

\bibitem[Fumagalli et~al.(2025)Fumagalli, Muschalik, H{\"u}llermeier, Hammer, and Herbinger]{shapley_coopGames}
Fabian Fumagalli, Maximilian Muschalik, Eyke H{\"u}llermeier, Barbara Hammer, and Julia Herbinger.
\newblock Unifying feature-based explanations with functional anova and cooperative game theory.
\newblock In \emph{The 28th International Conference on Artificial Intelligence and Statistics}, 2025.

\bibitem[Mohammadi et~al.(2025{\natexlab{a}})Mohammadi, Chau, and Muandet]{mohammadi2025computing}
Majid Mohammadi, Siu~Lun Chau, and Krikamol Muandet.
\newblock Computing exact shapley values in polynomial time for product-kernel methods.
\newblock \emph{arXiv preprint arXiv:2505.16516}, 2025{\natexlab{a}}.

\bibitem[Shapley(1953)]{shapley}
Lloyd~S Shapley.
\newblock A value for n-person games.
\newblock \emph{Contributions to the Theory of Games}, 2\penalty0 (28):\penalty0 307--317, 1953.

\bibitem[Ma et~al.(2008)Ma, Gao, Li, Jiang, Guo, et~al.]{ma2008shapley}
Ying Ma, Zuofeng Gao, Wei Li, Ning Jiang, Lei Guo, et~al.
\newblock The shapley value for stochastic cooperative game.
\newblock \emph{Modern Applied Science}, 2\penalty0 (4):\penalty0 1--76, 2008.

\bibitem[Duvenaud et~al.(2011)Duvenaud, Nickisch, and Rasmussen]{AGP}
David~K Duvenaud, Hannes Nickisch, and Carl~E Rasmussen.
\newblock Additive gaussian processes.
\newblock In \emph{Advances in neural information processing systems}, pages 226--234, 2011.

\bibitem[Mohammadi et~al.(2025{\natexlab{b}})Mohammadi, Tiddi, and Ten~Teije]{gemfix}
Majid Mohammadi, Ilaria Tiddi, and Annette Ten~Teije.
\newblock Unlocking the game: Estimating games in m{\"o}bius representation for explanation and high-order interaction detection.
\newblock In \emph{Proceedings of the AAAI Conference on Artificial Intelligence}, volume~39, pages 19512--19519, 2025{\natexlab{b}}.

\bibitem[Mohammadi et~al.(2025{\natexlab{c}})Mohammadi, Tiddi, and Ten~Teije]{svsvl}
Majid Mohammadi, Ilaria Tiddi, and Annette Ten~Teije.
\newblock Support vector-based estimation of multilinear games for feature selection and explanation.
\newblock In \emph{Proceedings of the AAAI Conference on Artificial Intelligence}, volume~39, pages 19520--19527, 2025{\natexlab{c}}.

\bibitem[Chau et~al.(2021{\natexlab{a}})Chau, Ton, González, Teh, and Sejdinovic]{chau_bayesimp_2021}
Siu~Lun Chau, Jean-Francois Ton, Javier González, Yee Teh, and Dino Sejdinovic.
\newblock {BayesIMP}: {Uncertainty} {Quantification} for {Causal} {Data} {Fusion}.
\newblock In \emph{Advances in {Neural} {Information} {Processing} {Systems}}, volume~34, pages 3466--3477. Curran Associates, Inc., 2021{\natexlab{a}}.

\bibitem[Chau et~al.(2021{\natexlab{b}})Chau, Bouabid, and Sejdinovic]{chau_deconditional_2021}
Siu~Lun Chau, Shahine Bouabid, and Dino Sejdinovic.
\newblock Deconditional {Downscaling} with {Gaussian} {Processes}.
\newblock In \emph{Advances in {Neural} {Information} {Processing} {Systems}}, volume~34, pages 17813--17825. Curran Associates, Inc., 2021{\natexlab{b}}.

\bibitem[Chau et~al.(2022)Chau, Hu, Gonzalez, and Sejdinovic]{chau2022rkhs}
Siu~Lun Chau, Robert Hu, Javier Gonzalez, and Dino Sejdinovic.
\newblock Rkhs-shap: Shapley values for kernel methods.
\newblock \emph{Advances in neural information processing systems}, 35:\penalty0 13050--13063, 2022.

\bibitem[Adachi et~al.(2024)Adachi, Planden, Howey, Osborne, Orbell, Ares, Muandet, and Chau]{adachi2024looping}
Masaki Adachi, Brady Planden, David Howey, Michael~A Osborne, Sebastian Orbell, Natalia Ares, Krikamol Muandet, and Siu~Lun Chau.
\newblock Looping in the human: Collaborative and explainable bayesian optimization.
\newblock In \emph{International Conference on Artificial Intelligence and Statistics}, pages 505--513. PMLR, 2024.

\bibitem[{\v{S}}trumbelj and Kononenko(2014)]{sampling_shap}
Erik {\v{S}}trumbelj and Igor Kononenko.
\newblock Explaining prediction models and individual predictions with feature contributions.
\newblock \emph{Knowledge and information systems}, 41:\penalty0 647--665, 2014.

\bibitem[Covert and Lee(2021)]{unbiased_shap}
Ian Covert and Su-In Lee.
\newblock Improving kernelshap: Practical shapley value estimation using linear regression.
\newblock In \emph{International Conference on Artificial Intelligence and Statistics}, pages 3457--3465. PMLR, 2021.

\bibitem[Masoomi et~al.(2021)Masoomi, Hill, Xu, Hersh, Silverman, Castaldi, Ioannidis, and Dy]{bivariateSHAP}
Aria Masoomi, Davin Hill, Zhonghui Xu, Craig~P Hersh, Edwin~K Silverman, Peter~J Castaldi, Stratis Ioannidis, and Jennifer Dy.
\newblock Explanations of black-box models based on directional feature interactions.
\newblock In \emph{International Conference on Learning Representations}, 2021.

\bibitem[Plumb et~al.(2018)Plumb, Molitor, and Talwalkar]{maple}
Gregory Plumb, Denali Molitor, and Ameet~S Talwalkar.
\newblock Model agnostic supervised local explanations.
\newblock \emph{Advances in neural information processing systems}, 31, 2018.

\bibitem[Climente-Gonz{\'a}lez et~al.(2019)Climente-Gonz{\'a}lez, Azencott, Kaski, and Yamada]{hsic_lasso}
H{\'e}ctor Climente-Gonz{\'a}lez, Chlo{\'e}-Agathe Azencott, Samuel Kaski, and Makoto Yamada.
\newblock Block hsic lasso: model-free biomarker detection for ultra-high dimensional data.
\newblock \emph{Bioinformatics}, 35\penalty0 (14):\penalty0 i427--i435, 2019.

\bibitem[Vergara and Est{\'e}vez(2014)]{mi_fs}
Jorge~R Vergara and Pablo~A Est{\'e}vez.
\newblock A review of feature selection methods based on mutual information.
\newblock \emph{Neural computing and applications}, 24:\penalty0 175--186, 2014.

\bibitem[Tibshirani(1996)]{lasso}
Robert Tibshirani.
\newblock Regression shrinkage and selection via the lasso.
\newblock \emph{Journal of the Royal Statistical Society Series B: Statistical Methodology}, 58\penalty0 (1):\penalty0 267--288, 1996.

\bibitem[Geurts et~al.(2006)Geurts, Ernst, and Wehenkel]{tree_ensemble}
Pierre Geurts, Damien Ernst, and Louis Wehenkel.
\newblock Extremely randomized trees.
\newblock \emph{Machine learning}, 63:\penalty0 3--42, 2006.

\end{thebibliography}
